%% file: main.tex
\pdfoutput=1
\documentclass[10pt,onecolumn,a4paper]{fairmeta}

\input{preamble}

\newcommand{\papertitle}{SSDD: Single-Step Diffusion Decoder\\for Efficient Image Tokenization}

\title{\papertitle}
\author[1,2]{Théophane Vallaeys}
\author[1]{Jakob Verbeek}
\author[2,3]{Matthieu Cord}

\affiliation[1]{Meta Fundamental AI Research}
\affiliation[2]{Sorbonne University}
\affiliation[3]{Valeo.ai}

\abstract{\input{parts/abs}}

\date{October 7, 2025 (Updated: June 29, 2026)}
\correspondence{Théophane Vallaeys (\email{webalorn@meta.com}) }
\metadata[Code]{\url{https://github.com/facebookresearch/SSDD}}

\begin{document}

\maketitle

\input{parts/intro}

\input{parts/related}
\input{parts/method}

\input{parts/exp}

\input{parts/concl}

\bibliographystyle{assets/plainnat}
\bibliography{bibliography}

\input{parts/supmat}

\end{document}

%% file: preamble.tex
\usepackage{multirow}
\usepackage{tabularx}
\usepackage{makecell}
\usepackage{stmaryrd}

\usepackage{xspace}
\usepackage{amsmath,amsfonts,bm,amsthm}
\usepackage{amssymb}
\usepackage{mathtools}
\usepackage{sidecap}
\usepackage{enumitem}
\usepackage{soul}
\usepackage{array}

\usepackage{caption}
\usepackage{subcaption}
\usepackage{ifthen}
\usepackage{tikz}
\usetikzlibrary{arrows.meta,arrows,tikzmark}
\usepackage{float}
\usepackage{microtype}
\usepackage{graphicx}

\usepackage{booktabs}
\usepackage{wrapfig}
\usepackage{orcidlink}
\usepackage{marvosym}

\usepackage{rotating}
\usepackage{multirow}
\usepackage[table,dvipsnames]{xcolor}

\makeatletter
\DeclareRobustCommand\onedot{\futurelet\@let@token\@onedot}
\def\@onedot{\ifx\@let@token.\else.\null\fi\xspace}
\makeatother

\def\ie{\emph{i.e}\onedot}

\def\oinc{$\uparrow$}
\def\odec{$\downarrow$}

\def\upsymb#1{\ensuremath{\ifcase#1\or *\or \dagger\or \ddagger\or
   \mathsection\or \mathparagraph\or \|\or **\or \dagger\dagger
   \or \ddagger\ddagger \else\@ctrerr\fi}}

\usepackage{cleveref}
\crefname{section}{Sec.}{Secs.}
\Crefname{section}{Section}{Sections}
\crefname{table}{Tab.}{Tabs.}
\Crefname{table}{Table}{Tables}
\crefname{figure}{Fig.}{Figs.}
\Crefname{figure}{Figure}{Figures}
\crefname{appendix}{App.}{App.}
\Crefname{appendix}{Appendix}{App.}
\crefname{equation}{Eq.}{Eqs.}

\def\mypar#1{\vspace{2mm}\noindent\textbf{#1.}\hspace{2mm}}
\newcolumntype{H}{>{\setbox0=\hbox\bgroup}c<{\egroup}@{}}
\def\res#1{$#1\!\times\!#1$}

\def\tabfigendspace{{\vspace{0mm}}}
\def\figmidspace{{\vspace{0mm}}}

\newcommand{\E}{\mathbb{E}}

\newcommand{\R}{\mathbb{R}}

\newcommand{\cl}{\mathcal{L}}

\newcommand{\cn}{\mathcal{N}}

\DeclareMathOperator{\DSample}{Sample}
\DeclareMathOperator{\LPIPSFN}{LPIPS}
\DeclareMathOperator{\REPAFN}{REPA}
\DeclareMathOperator{\FMFN}{FM}
\DeclareMathOperator{\fnsign}{sign}

\def\Ssdd{\textsc{SSDD}\xspace}
\def\ssddone{\Ssdd}
\def\ssddmulti{\textsc{SSDD}}
\def\sdvae{\textsc{SD-VAE}\xspace}
\def\Dito{\textsc{DiTo}\xspace}
\def\evae{$\varepsilon$-\textsc{VAE}\xspace}

\definecolor{cyellow}{HTML}{fff2cc}
\definecolor{cblue}{HTML}{cfe2f3}
\definecolor{cgreen}{HTML}{d9ead3}
\definecolor{cpink}{HTML}{f4d4e2}
\definecolor{cpurple}{HTML}{bbb0d4}
\definecolor{corange}{HTML}{fde4cd}
\definecolor{cred}{HTML}{e6b8af}

\def\rowStart{\addlinespace[0.5pt]}
\def\rowS{\rowStart}
\def\rowB{\rowStart\rowcolor{cgreen}}
\def\rowM{\rowStart\rowcolor{cblue}}
\def\rowL{\rowStart\rowcolor{cyellow}}
\def\rowXL{\rowStart\rowcolor{corange}}
\def\rowH{\rowStart\rowcolor{cpink}}
\def\rowXH{\rowStart\rowcolor{cred}}

%% file: parts/abs.tex
Tokenizers are a key component of state-of-the-art generative image models, extracting the most important features from the signal while reducing data dimension and redundancy.
Most current image tokenizers are based on KL-regularized variational autoencoders (KL-VAE), trained with reconstruction, perceptual and adversarial losses.
Diffusion decoders have been proposed as a more principled alternative to model the distribution over images conditioned on the latent. 
However, matching the performance of KL-VAE still required 
adversarial losses, as well as a higher decoding time due to  iterative sampling. 
To address these limitations, we introduce a new  pixel diffusion decoder architecture for improved scaling and training stability, benefiting from transformer components and GAN-free training.
We  use distillation to replicate the performance of the diffusion decoder in an efficient single-step decoder.
This makes \Ssdd{} the first GAN-free single-step diffusion decoder to reach state-of-the-art continuous tokenization, with higher reconstruction quality and faster sampling than KL-VAE.
In particular, \Ssdd{} improves reconstruction FID from $0.87$ to $0.46$ with $1.4\times$ higher throughput and preserves generation quality of DiTs with $3.8\times$ faster sampling.
As such, \Ssdd{} can be used as a drop-in replacement for KL-VAE, and for building higher-quality and faster generative models.

%% file: parts/intro.tex
\section{Introduction}
\label{sec:intro}

Current state-of-the-art image generation models, whether they are based on diffusion~\citep{sd1_Rombach_2022_CVPR}, flow matching~\citep{sd3_pmlr-v235-esser24a} or autoregressive modeling~\citep{vqgan_taming_transformers_high_res,var_NEURIPS2024_9a24e284},
rely on tokenizers as a key component to reduce the high-dimensional visual signal to compact latent representations.
This allows for efficient training and inference of these models in the latent space, before decoding the generated latent representations back to the original signal representation in pixel-space.
Beyond efficiency, tokenization has also been shown to improve generative performance~\citep{sd1_Rombach_2022_CVPR,vqgan_taming_transformers_high_res}.
The quality and compression ratio of the tokenizer directly affect the generative quality of latent models.
Image tokenization models either produce discrete tokens ---directly compatible with, e.g., discrete text models such as LLMs---, or continuous tokens ---which  can be modeled with methods such as diffusion and flow matching.
In this work we focus on \textit{continuous} tokenization, as used in  most state-of-the-art models for images and videos.

\input{parts/figures/intro}

For generative modeling, tokenizers are usually trained as autoencoders.
They use an encoder $E: x \mapsto z$ to compress an image $x\in\R^{3\times H\times W}$ into a latent representation $z\in\R^{c\times \frac{H}{f}\times \frac{W}{f}}$, where $f$ is the spatial downsampling factor and $c$ is the channel dimension of the latent.
The  decoder $D: z \mapsto \hat{x}$ recovers an estimate $\hat{x}$ of the original signal $x$.
As the encoding process is lossy when using typical configurations with $3f^2> c$,
the decoding process can be seen as a generative task conditioned on the latent representation $z=E(x)$.
Reconstruction quality can be quantified in different ways:
\begin{enumerate}
    \item 
 \textbf{Distortion:}
    expressed as $\E_{x}\Delta(x,\hat{x})$, where $\hat{x}=D(E(x))$, and $\Delta$ is a  measure of the distortion \emph{between the original and  reconstructed  images}.
    Distortion can be quantified by low-level similarity metrics such as MSE, MAE, PSNR and SSIM, or by high-level metrics based on features found in trained deep neural networks, such as LPIPS~\citep{lpips_Zhang_2018_CVPR} and DreamSim~\citep{dreamsim_fu23arxiv}.
\item \textbf{Distribution shift:}
    expressed as $d(P_x,P_{\hat{x}})$, with $d$ being a measure of the divergence \emph{between the original and  reconstructed image distributions}.
    Metrics commonly used to evaluate distribution shift of generative models include Fréchet Inception Distance (FID)~\citep{FID_metric_NIPS2017_8a1d6947},
    as well as Density and Coverage~\citep{density_coverage_pmlr-v119-naeem20a}.
\end{enumerate}
As shown by Blau and Michaeli~\citep{Blau_2018_CVPR}, there exists a fundamental trade-off between these two different notions of reconstruction quality (referred to as \textit{Distortion} and \textit{Perception} in \citet{Blau_2018_CVPR})   when the signal is compressed sufficiently. 
While Distortion can be minimized by a deterministic decoder, Distribution shift is  optimized using a generative decoder.
See \Cref{app:theory} for more detail and an illustrative example.

Common tokenizers such as the KL-VAE~\citep{sd1_Rombach_2022_CVPR} are trained using an L1 reconstruction loss, LPIPS~\citep{lpips_Zhang_2018_CVPR}, and a GAN discriminator~\citep{goodfellow14nips}, with an added KL-regularization on the latent space.
L1 and LPIPS losses directly optimize for distortion,
while the GAN loss aims to decrease distribution shift.
However, the diversity of these deterministic decoders, which affects the distribution shift, is limited by the expressivity of the latent $z$. 
In particular, for a given image $x$ and its encoding $z=E(x)$ there is only a single decoding possible.
Additionally, stable scaling of  GAN losses  is non-trivial~\citep{biggan_brock2018large}.
Diffusion auto-encoders \citep{dito_chen2025diffusionautoencodersscalableimage,evae_zhao2025epsilonvaedenoisingvisualdecoding}
have recently been proposed as an alternative to  directly model the conditional data distribution and optimizing for distributional shift.
Diffusion decoders naturally model the distribution of missing signal from the latents~\citep{swycc_birodkar2024samplecompress}, by being able to sample multiple decodings for a given latent, allowing higher compression ratios and focusing the task of latent modeling on higher-level features.
While these models have shown improved reconstruction and generation results at a given model size, they still suffer from (i) slow sampling, due to the iterative denoising process and (ii) the reliance on a GAN loss to achieve state-of-the-art results.

 In this paper we introduce \Ssdd{}, a new diffusion autoencoder that overcomes the limitations of previous diffusion decoders.  
In particular, we propose an efficient and scalable alternative to the simple convolutional U-Net architectures for diffusion decoders, based on the U-ViT~\citep{hoogeboom23simple,sid2_hoogeboom2025simplerdiffusionsid215} and achieve superior results at all tested model scales.
We experimentally analyze the key  design choices to train our model, which is the first GAN-free single-step diffusion decoder to reach state-of-the-art continuous tokenization, improving the scaling and stability of the training without any diverging runs across model scales.
To avoid multiple denoising steps while decoding,  we propose a single-step distillation method for \Ssdd{} with minimal quality impact.
See \Cref{fig:main_fig} for an illustration of the \Ssdd{} decoder training and sample results.

Taken together, our contributions make \Ssdd{} both the highest quality and fastest diffusion decoder, surpassing KL-VAE and previous diffusion auto-encoders on quality and reconstruction speed at the same downsampling factor.
In the f8c4 setting (8$\times$ downsampling, 4 latent channels) our smallest \Ssdd{}-S improves reconstruction FID over KL-VAE from $0.87$ to $0.46$ with $1.4\times$ higher decoding throughput. At matched throughput, \Ssdd{}-B reaches $0.42$ rFID. At matched parameter count, \Ssdd{}-M reaches $0.39$ rFID. As our architecture differs substantially from KL-VAE, decoding time is the most meaningful efficiency metric, and \Ssdd{} improves on all fronts.
By leveraging an SSDD auto-encoder with higher compression rate (f16c4), we increase the DiT sampling speed of latents by $3.8\times$ while preserving sample quality when decoded using our model.

%% file: parts/figures/intro.tex
\begin{figure}[t]
    \centering
    \includegraphics[width=.51\textwidth,trim={0 1.2em 0 1em},clip]{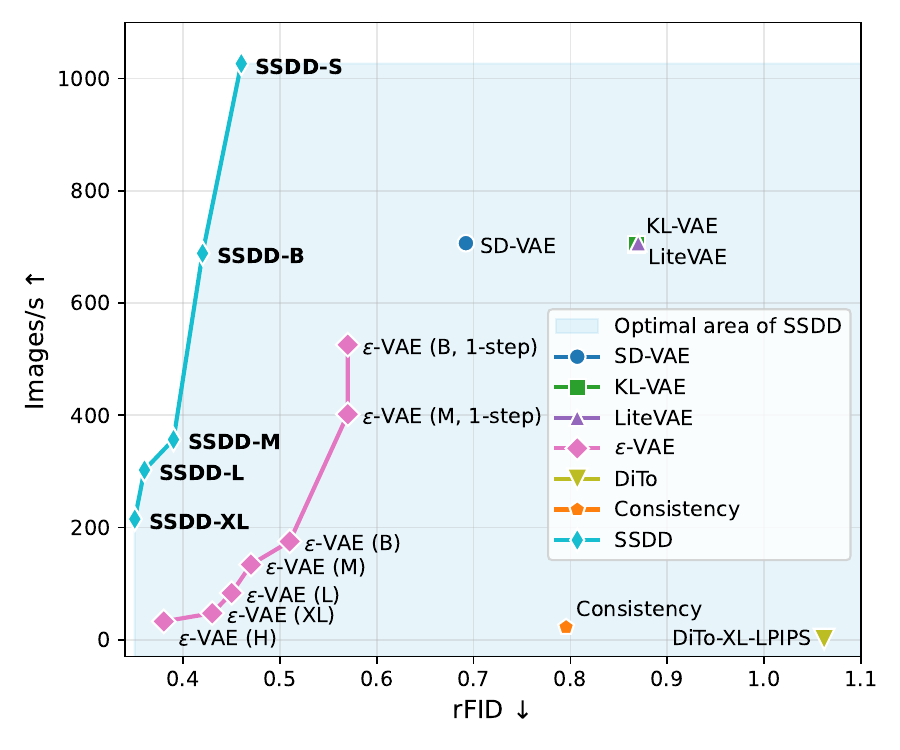}
    \hfill
    \includegraphics[width=.4\textwidth,trim={0 0.8em 0 0.6em},clip]{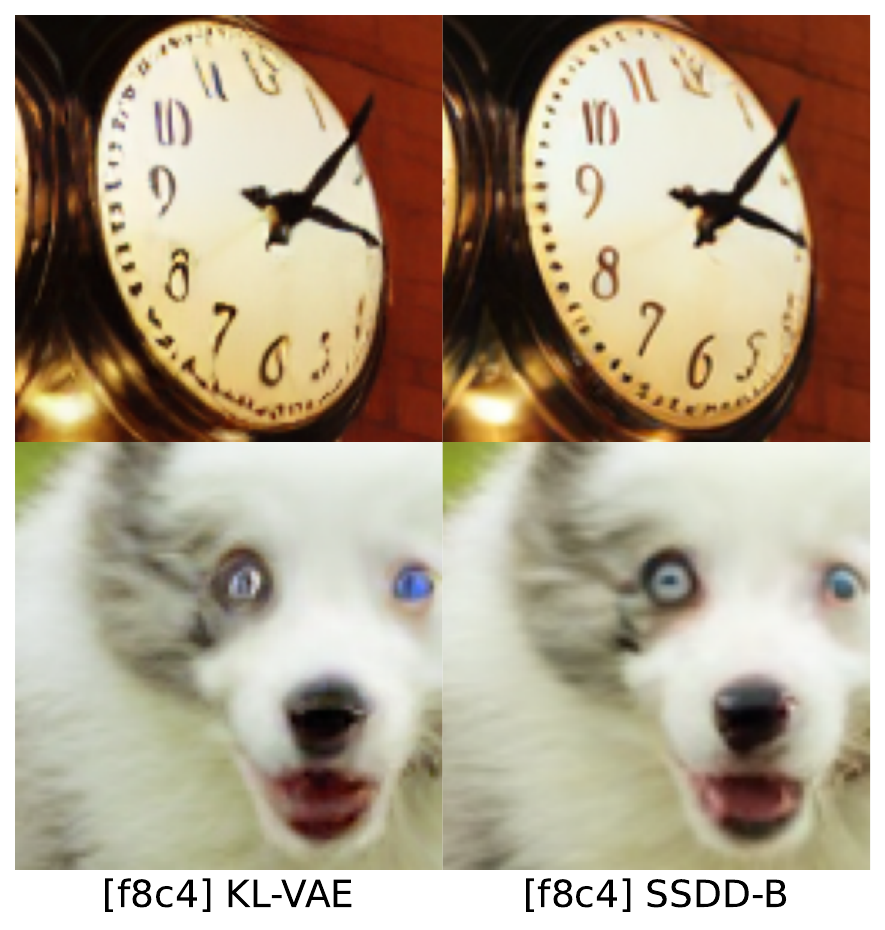}
    \includegraphics[width=0.9\textwidth,trim={0 1.3em 0 0.5em},clip]{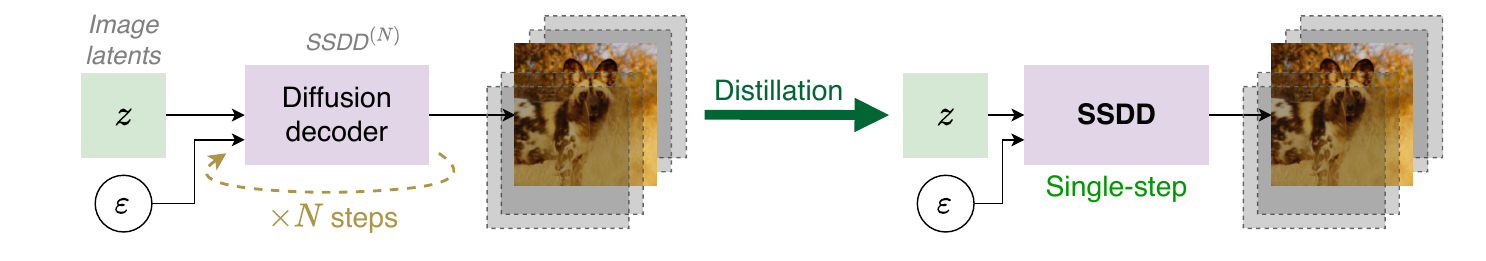}
    \caption{
        \textbf{Left:} Decoding speed-quality Pareto-front for different state-of-the-art f8c4 feedforward and diffusion decoders.
        \textbf{Right:} Zoomed-in patches from reconstructions obained with  KL-VAE and our SSDD tokenizer with similar throughput.
        \textbf{Bottom:} High-level overview of our SSDD decoder.
        First we train a multi-step decoder, generating diverse high-quality samples, and then distill it into a single-step decoder.
        Both multi-step and distilled single-step decoders can generate diverse decodings of a given latent $z$ by using the input noise $\varepsilon$ to sample from the distribution over images given the latent.
    }
    \label{fig:main_fig}
\end{figure}

%% file: parts/related.tex
\section{Related work}
\label{sec:related}

\mypar{Image tokenizers}
Tokenization is the first stage in  state-of-the-art generative  image models.
Tokenizers for diffusion models  compress images into continuous low-dimension representations~\citep{sd1_Rombach_2022_CVPR,dit_scalable_diffusion_Peebles_2023_ICCV}, using a convolutional autoencoder~\citep{autoencoder_doi:10.1126/science.1127647} trained with L1, KL, LPIPS~\citep{lpips_Zhang_2018_CVPR} and GAN~\citep{goodfellow14nips} losses, which we refer to as KL-VAE.
Discrete autoencoders are used as tokenizers for discrete autoregressive modeling~\citep{vqvae_NIPS2017_7a98af17,vqvae2_NEURIPS2019_5f8e2fa1,maskgit_Chang_2022_CVPR,magvit_Yu_2023_CVPR,improved_vqgan_yu2022vectorquantized,vqgan_taming_transformers_high_res,llamagen_sun2024autoregressivemodelbeatsdiffusion}, replacing the KL regularization with quantization.
Advances have been made on multiple aspects of those models, including efficiency~\citep{litevae_NEURIPS2024_0735ab35},
semantic alignment~\citep{vavae_Yao_2025_CVPR},
architecture and training~\citep{maetok_chen2025masked,l_detok_yang2025latentdenoisingmakesgood},
spatial downsampling~\citep{dcae_chen2025deep},
quantization methods~\citep{magvit2_yu2024language,zhao2025image,fsq_mentzer2024finite,Lee_2022_CVPR,pmlr-v202-huh23a},
and 1D-tokenization~\citep{titok_NEURIPS2024_e91bf7df,softvqvae_Chen_2025_CVPR}.
However, most tokenizers rely on  deterministic decoders  optimized for  image-to-image distortion, but may be suboptimal in terms of distribution shift~\citep{Blau_2018_CVPR}. 

\mypar{Diffusion decoders}
Diffusion decoders~\citep{dae_Preechakul_2022_CVPR,divae_shi2022divaephotorealisticimagessynthesis,soda_Hudson_2024_CVPR,dgae_liu2025dgaediffusionguidedautoencoderefficient,pandey2022diffusevae}, usually based on U-Net models~\citep{unet_2015}, are trained using a diffusion loss.
They have been applied in a compression setting~\citep{high_fidelity_image_compression_generative_models2023,swycc_birodkar2024samplecompress,NEURIPS2023_ccf6d8b4,careil24iclr} in place or in addition to deterministic decoders to improve quality by modeling the distribution of realistic reconstructions.
Some variants also use diffusion decoders as an additional stage within a KL-VAE~\citep{Wurstchen_2025,flextok_bachmann2025flextok}, further compressing the encoding space.
Recent work has shown that these models are scalable and easy to train when only using the diffusion loss~\citep{dito_chen2025diffusionautoencodersscalableimage}, and, at the cost of adding additional LPIPS and GAN losses, can be competitive to similarly-sized VAEs~\citep{evae_zhao2025epsilonvaedenoisingvisualdecoding,flow_to_the_mode_sargent2025flowmodemodeseekingdiffusion,dalle3}.
Diffusion decoders still suffer from slow multi-step sampling and the reliance on hard-to-scale adversarial training~\citep{biggan_brock2018large}.
Our work aims to alleviate these limitations.

\mypar{Diffusion and flow image modeling}
Diffusion and flow matching models estimate the reverse process of a forward process that interpolates between data and pure Gaussian noise~\citep{NEURIPS2019_3001ef25,NEURIPS2020_4c5bcfec,song2021scorebased}.
They were first shown by~\citet{diffusion_beats_gan_NEURIPS2021_49ad23d1} to outperform previous methods such as
GANs~\citep{goodfellow14nips} and VAEs~\citep{kingma2013auto}.
Diffusion  models, often operating in a compressed latent space for efficiency~\citep{sd1_Rombach_2022_CVPR},
 underlie  current state-of-the-art models for text-to-image~\citep{imagen_2024,glide_pmlr-v162-nichol22a,sd1_Rombach_2022_CVPR,sdxl_podell2024sdxl},
image-to-image~\citep{palette_2022},
and video generation~\citep{10.1007/978-3-031-72986-7_23,emu_video_2023,moviegen_polyak2024movie,nvidia2025cosmosworldfoundationmodel}.
Ongoing research efforts aim to improve various aspects of these models, including training, sampling speed and model architecture~\citep{NEURIPS2022_a98846e9,dit_scalable_diffusion_Peebles_2023_ICCV,diffusion_elbo_dataaug_NEURIPS2023_ce79fbf9,improved_ddpm_pmlr_v139_nichol21a,chen2023importancenoiseschedulingdiffusion}.
In particular, flow matching~\citep{flow_matching_lipman2023flow,flow_straight_liu2023flow,nf_si_albergo2023building} reformulates the diffusion as a velocity field estimation between noise and data, yielding improvements and faster sampling in generative image models~\citep{sd3_pmlr-v235-esser24a}.
Pixel-space diffusion has seen progress with new training architectural improvements~\citep{hoogeboom23simple,sid2_hoogeboom2025simplerdiffusionsid215,hourglass_diffusion_transformers_pmlr_v235_crowson24a}.
As an alternative to diffusion, autoregressive models have seen fast development both for discrete and continuous modeling~\citep{vqvae_NIPS2017_7a98af17,maskgit_Chang_2022_CVPR,vqgan_taming_transformers_high_res,var_NEURIPS2024_9a24e284,yu2024randomizedautoregressivevisualgeneration,chameleonteam2025chameleonmixedmodalearlyfusionfoundation,liang2025mixtureoftransformers}.
Recent approaches have also been mixing both methods to improve generation~\citep{diffusion_forcing_NEURIPS2024_2aee1c41,diffusion_loss_NEURIPS2024_66e22646,zhou2025transfusion,tang2025hart}.
Our work leverages continuous flow matching for generative modeling in pixel space.

\mypar{Few-step inference}
The main drawback of  diffusion models is the  iterative inference  process which requires multiple, often several tens or hundreds, forward passes through the model. 
To address this, few- or single-step sampling methods have been developed, through distillation or trajectory alignment.
Consistency models~\citep{consistency_pmlr-v202-song23a,improved_consistency_song2024improved,luo2023latentconsistencymodelssynthesizing} distill pre-trained diffusion model capabilities into new few-step tailored generator models. 
Other methods  fine-tune the diffusion model using a teacher-student setup for fast sampling~\citep{one_step_distrib_matching_Yin_2024_CVPR,improved_distribu_matching_NEURIPS2024_54dcf253,liu2024instaflow,luhman2021knowledgedistillationiterativegenerative,Meng_2023_CVPR,salimans2022progressive,NEURIPS2023_9c2aa1e4}.
We leverage this approach and study its impact on decoding performance of our models.

%% file: parts/method.tex
\section{Single-step diffusion decoder method}
\label{sec:method}

We introduce here the main components of our method \Ssdd{} combining several key innovations to make diffusion decoders both efficient and practical. First, we introduce a hybrid U-Net--transformer architecture that leverages convolutional inductive biases while scaling effectively with transformer blocks (\Cref{subsec:method_architecture}).  
Second, we propose a GAN-free training scheme based on flow matching, perceptual alignment, and feature regularization, ensuring stable training without adversarial losses (\Cref{subsec:method_training}). 
Third, we design a single-step distillation process that transfers the behavior of multi-step diffusion to a fast one-step decoder, achieving both quality and speed (\Cref{subsec:sampling_distill}). 
Finally, we show that \Ssdd{} can operate with shared encoders, enabling compatibility with existing VAEs and reducing training costs (\Cref{subsec:encoder}). Together, these components make \Ssdd{} the first diffusion decoder optimized for single-step reconstruction, without compromising generative performance.

\input{parts/figures/training_overview}

\subsection{A scalable pixel-space diffusion decoder architecture}
\label{subsec:method_architecture}

Existing pixel-space diffusion decoders use the convolutional U-Net architecture~\citep{evae_zhao2025epsilonvaedenoisingvisualdecoding} that was found to be successful in early pixel-space diffusion models~\citep{diffusion_beats_gan_NEURIPS2021_49ad23d1}.
Transformer architectures, however, have proven to be more effective and scalable for latent modeling~\citep{dit_scalable_diffusion_Peebles_2023_ICCV}.
We aim to combine the strengths of convolutional U-Nets to model features in pixel-space with the generative capabilities of diffusion transformers. 
We base our architecture on the U-ViT~\citep{hoogeboom23simple} with several modifications.
See \Cref{fig:model} for an overview.

\mypar{U-Net with latent transformer}
Based on the U-ViT architecture, we modify a U-Net~\citep{unet_2015,diffusion_beats_gan_NEURIPS2021_49ad23d1} using four levels of two convolutional ResNet blocks and three convolution downsampling / upsampling layers 
by removing all attention layers and replacing the middle attention block by a transformer operating on tokens representing $8\!\times\!8$ pixel patches.
Each transformer block uses the GEGLU activation function~\citep{geglu_shazeer2020gluvariantsimprovetransformer}, a $4\times$ multiplicative factor for the hidden MLP dimension, and multi-head attention~\citep{attention_is_all_you_need_NIPS2017_3f5ee243}.

\mypar{Time and position embedding}
We embed time  using adaptive group-normalization (AdaGN) \citep{diffusion_beats_gan_NEURIPS2021_49ad23d1} as the second normalization layer in ResNet blocks.
To enforce a local inductive bias,
we learn a relative positional embedding table~\citep{relative_pos_shaw2018selfattentionrelativepositionrepresentations} for each self-attention layer,
with visual tokens only attending to tokens up to a distance of $8$ on the width and height axes.
This results in a fixed $17\!\times\!17$ attention window and relative positional embedding table.

\mypar{Conditioning}
Following DiTo~\citep{dito_chen2025diffusionautoencodersscalableimage} and \evae{}~\citep{evae_zhao2025epsilonvaedenoisingvisualdecoding}, we condition our model by upsampling the latent feature grid $z$ from $(c,\frac{H}{f},\frac{W}{f})$ to $(c,H,W)$, and concatenating it with the noised image along the channel dimension in our model.
As our model is relatively deep and narrow compared to prior work, we improve conditioning by adding a second mechanism using adaptive normalization.
We replace the first GroupNorm of ResNet blocks with AdaGN,
and the first LayerNorm of transformer blocks by AdaLN \citep{dit_scalable_diffusion_Peebles_2023_ICCV}.

\mypar{Scaling}
To study a wide range of decoding capacities, we scale our diffusion decoders from 13.4M (\Ssdd{}-S) to 153.8M parameters (\Ssdd{}-XL) by increasing the number of channels and of transformer layers.
For most of our experiments we consider \Ssdd{}-M (48M parameters), which has a similar size as the convolutional decoder of  KL-VAE~\citep{sd1_Rombach_2022_CVPR} (47.2M parameters).
More details about model parametrization can be found in \Cref{app:implementation_details}.

\subsection{GAN-free training}
\label{subsec:method_training}

\mypar{Flow matching}
We train our model using the optimal transport flow-matching loss~\citep{flow_matching_lipman2023flow}  with $\sigma_{min}=0$.
Specifically, we use $x_t=(1-t)x + t\varepsilon$, and our decoder $D(x_t|t,z)$ is trained to predict the velocity field $\nu=x-\varepsilon$ with the L2 loss.
This defines our main training loss $\cl_{\FMFN}=\parallel x - \varepsilon - D(x_t|t,z)\parallel^2$.
Following \citet{sd3_pmlr-v235-esser24a}, we sample $t$ during training using the logit-normal distribution \citep{atchison1980logistic} with location $m=0$ and scale $s=1$.

\mypar{Perceptual and regularization losses}
As noted by \citet{dito_chen2025diffusionautoencodersscalableimage}, perceptual loss is important to guide the generation toward perceptually correct areas.
We therefore add the LPIPS \citep{lpips_Zhang_2018_CVPR} loss $\cl_{\LPIPSFN}=\LPIPSFN(x,\hat{x}_0)$, with $\hat{x}_0=x_t + t D(x_t|t,z)$ the single-step prediction of $x_0$ from $x_t$ and $z$.
As the middle part of our model acts as an implicit latent-space diffusion transformer, we further stabilize and accelerate our training by adding the REPA loss~\citep{repa_yu2025representation}, denoted as $\cl_{\REPAFN}$.
We use  DINOv2-B~\citep{dinov2_oquab2024dinov}  reference features for REPA, which are aligned with the tokens extracted from the 4th layer of our transformer, through a 2-layer MLP.
The FM loss trains the decoder to predict the clean image trajectory, while LPIPS aligns perceptual features and REPA stabilizes transformer representations.
We combine them in our final loss as $\cl=\cl_{\FMFN} + \lambda_{\LPIPSFN}\cl_{\LPIPSFN} + \lambda_{\REPAFN}\cl_{\REPAFN}$, with $\lambda_{\LPIPSFN}=0.5$ and $\lambda_{\REPAFN}=0.25$.

\mypar{Multi-scale adaptation}
While autoencoders have been observed to generalize to higher resolutions than the one used for training~\citep{litevae_NEURIPS2024_0735ab35}, best performance is often reached when training or fine-tuning at specific higher target resolutions.
To alleviate the need of training from scratch at multiple resolutions and reduce the required training compute, we use a two-stage training method, based on the procedure from \citet{litevae_NEURIPS2024_0735ab35}.
During the first stage, we train a model on $128\!\times\!128$ crops.
We additionally reduce the divergence between the features distribution of the training and fine-tuning images by adding random resizing to the data augmentation, from which we extract a $128\!\times\!128$ crop.
During the second and third stages, we fine-tune and distill the first-stage weights at the target resolution (either $128\!\times\!128$ or $256\!\times\!256$).
Given the increased cost of training at higher resolutions, we use the second model (optimized for $256\times 256$) to evaluate on larger images (e.g. $512\times 512$ or $1024\times 1024$), which we observe to generalize sufficiently well.

\subsection{Single-step diffusion decoder method}
\label{subsec:sampling_distill}

A key limitation of diffusion decoders is their iterative nature, which limits throughput in downstream generative models. We propose to align the behavior of a multi-step diffusion decoder with the one of a single-step generator, without sacrificing quality.

\mypar{Sampling behavior and non-straight velocity dynamics}
We first analyze the effect of the number of sampling steps $N$. 
As previously observed by \citet{evae_zhao2025epsilonvaedenoisingvisualdecoding}, reconstruction quality is non-monotonic: we find that one-step denoising yields the best low-level distortion metrics (e.g., PSNR), while perceptual metrics (such as LPIPS and rFID) improve up to an intermediate number of steps before degrading. See \Cref{app:sampling_distill} for corresponding experimental results quantifying these effects.

This behavior is rooted in the interaction between the flow-matching objective and the LPIPS loss. Let $x_0$ be the original image, $\nu = x_0 - \varepsilon$ the true velocity, and $\hat{\nu} = D(x_t, z)$ the predicted velocity. We define the composite reconstruction loss as $$\mathcal{L}_\lambda(\hat{\nu}|\nu) = \Vert\hat{\nu}-\nu\Vert^2 + \lambda L(\hat{x}_0, x_0),$$ where $L$ is the LPIPS loss and $\hat{x}_0 = x_t + t\hat{\nu}$. The gradient of this loss with respect to $\hat{\nu}$ is:
\begin{align}
    \nabla_{\hat\nu}\cl_\lambda &= 2(\hat{\nu} - \nu) + \lambda t \nabla L \\
    &=2(\hat{\nu} - (\nu - \frac{\lambda t}{2} \nabla L)) \\
    &=\nabla_{\hat\nu}\cl_0(\hat\nu|\nu - \frac{\lambda t}{2} \nabla L).
\end{align}
This reveals that optimizing $\mathcal{L}_\lambda$ is equivalent to learning a velocity field $\nu - \frac{\lambda t}{2} \nabla L$, which is \textit{no longer straight}: $(\nabla_{\hat\nu}L)L(\hat{x}_0,x_0)$ is not necessarily aligned with $\nu$, and its impact increases with the noise level $t$. Consequently, LPIPS-regularized Flow-Matching decoders learn non-straight trajectories that shift during training. While this hinders denoising with many steps, it provides a ``sweet spot'' for high-quality, few-step reconstruction. 
We discuss this behavior further in \Cref{app:sampling_distill}. As we target diversity and quality for generative applications over reconstruction fidelity, we start with large denoising steps before adding a few denoising steps to refine details.

We adopt the scheduler from \citet{flow_to_the_mode_sargent2025flowmodemodeseekingdiffusion}, yielding the timesteps schedule $t_i = (\frac{N-i+1}{N})^\rho$ with $\rho=2$ and $N=8$ steps for all our \Ssdd{} models, to balance realism (\textit{low distributional shift}) and fidelity (\textit{low distortion}).

\mypar{One-step sampling behavior alignment by distillation}
To address the key limitations of diffusion decoders, \ie their slow multi-step inference process, we align the behavior of a single-step generation with that of multi-step through a lightweight distillation strategy.
Specifically, we use a frozen copy of our decoder $D_\text{ref}$ as a teacher to produce multi-step reconstructions $\hat{x}_\text{ref}$, and fine-tune the student decoder $D$ to reproduce this behavior in a single step.
Unlike~\citet{luhman2021knowledgedistillationiterativegenerative} that  solely rely on an L2 regression loss, we preserve the full training objective during distillation, with flow-matching and LPIPS terms being computed against teacher outputs. 
Each iteration, we sample $z\sim E(x)$, $\varepsilon\sim\cn(0;I)$,  $\hat{x}_\text{ref}=\DSample_{\text{steps}=N}(D_\text{ref},\varepsilon,z)$.

We fine-tune $D$ using $t\!=\!1$, the same $z, \varepsilon$ values and the altered loss $\cl^\text{distill}$,
with $\cl^\text{distill}_{\FMFN}=\parallel \hat{x}_\text{ref} - \varepsilon - D(x_t|t,z)\parallel^2$ and  $\cl^\text{distill}_{\LPIPSFN}=\LPIPSFN(x,\hat{x}_0)$. 
This allows the distilled decoder to mimic the perceptual and generative properties of the teacher, while operating in a single step and thus achieving much higher efficiency.
To our knowledge, this is the first work showing that the behavior of multi-step diffusion decoders can be aligned to a single-step generator without a strong quality drop, making \Ssdd{}  directly usable in generative pipelines.

\subsection{Shared encoders}
\label{subsec:encoder}

As our focus is on the decoder design, we use the standard KL-regularized convolutional  design from~\citet{sd1_Rombach_2022_CVPR} for the encoder.
We refer to encoders by their  patch size $f$ controlling the spatial downsampling, and output channel dimension $c$. 
Unless specified otherwise, we use the f8c4 encoder in  our experiments --- compressing RGB image patches of size $8\!\times\!8$ into  $4$ channels, achieving a compression factor of $8\times 8\times 3 /4 = 48$. 

\mypar{Shared encoder}
Previous works exploring diffusion autoencoders train a specific   encoder with each decoder. 
While in principle this allows to reach optimal results, it creates a distinct latent space associated  with each encoder.
In particular, this requires training a separate generative diffusion model associated with each model. 
Despite the encoder performance being dependent on the image resolution, we find that we can train a near-optimal multi-resolution encoder with a simple procedure.
We train an encoder together with a \Ssdd{}-M decoder, benefiting from multi-scale data augmentation, learning to encode features at different resolutions.
We then freeze this encoder and use it to train  decoders with a different number of parameters.
Unless specified otherwise, our decoders all rely on a single shared encoder for each combination of downsampling factor ($f$) and  channel count ($c$).

\mypar{Pretrained encoders}
\Ssdd{} can also be directly trained to decode images from any existing encoder mapping an input image to a grid-shaped latent representation.
in additional experiments in \Cref{app:additional_results} we explore this using various pretrained encoders, including
KL-VAE~\citep{sd1_Rombach_2022_CVPR}, DiTo~\citep{dito_chen2025diffusionautoencodersscalableimage}, and DC-AE~\citep{dcae_chen2025deep}.

%% file: parts/figures/training_overview.tex
\begin{figure}[t]
    \centering
    \includegraphics[width=\textwidth,trim={0 0.78em 0 0.5em},clip]{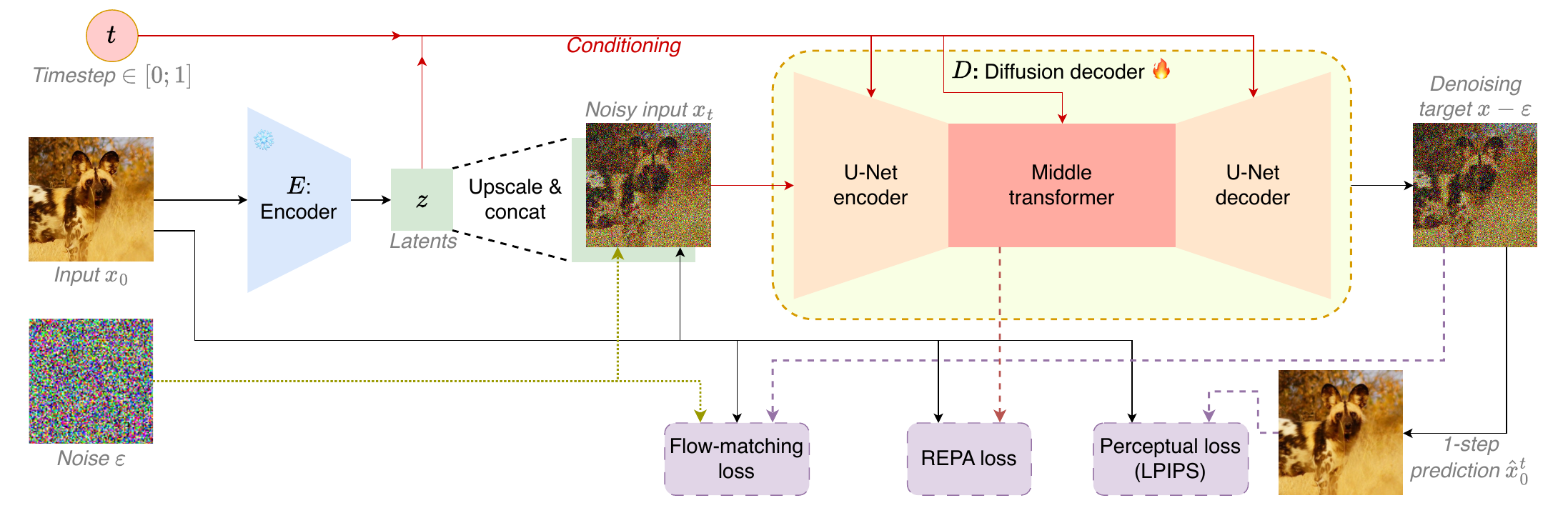}
    \caption{
        {\bf Training of \Ssdd{} tokenizer.} 
        Input image $x_0$ is mapped to latents $z$ by the (possibly frozen) encoder $E$.
        Noise $\epsilon\sim \mathcal{N}(0,1)$ is sampled and added to $x_0$ to form the noisy input $x_t$. 
        The decoder $D$ learns to denoise $x_t$  conditioned on denoising time $t$ (via AdaNorm) and latent $z$ (by using it as input to the decoder and  AdaNorm layers).
        The model is trained with flow-matching generative loss, REPA features alignment loss,  and LPIPS perceptual loss.
    }
    \label{fig:model}
\end{figure}

%% file: parts/exp.tex
\section{Experimental validation}
\label{sec:experiments}

\subsection{Experimental setup}
\label{subsec:setup}

We train and evaluate \Ssdd{} models of various sizes from \Ssdd{}-S to {\Ssdd{}-H} on ImageNet-1k~\citep{deng09cvpr}, and consider generalization of results to COCO~\citep{coco}.
For  details on model parametrization and training, see~\Cref{app:implementation_details}.
We use \ssddmulti$^{(N)}$ to refer to models using $N$ sampling steps, and plain \ssddone{} for single-step distilled models.
\looseness=-1

\mypar{Evaluation}
We evaluate the quality of our models on the 50k images of the ImageNet evaluation set.
For reconstruction, we assess the \textit{distribution shift} using the reconstruction Fréchet Inception Distance~(rFID)~\citep{FID_metric_NIPS2017_8a1d6947}.
We additionally evaluate the following distortion metrics: PSNR and SSIM~\citep{ssim_1284395} for low-level similarity, and LPIPS~\citep{lpips_Zhang_2018_CVPR} and DreamSim~\citep{dreamsim_fu23arxiv} as high-level perceptual metrics.
To evaluate the impact of our model on generation quality, we train Diffusion Transformer models~\citep{dit_scalable_diffusion_Peebles_2023_ICCV} on ImageNet-1k at $256\times256$ resolution using the original model configuration.
Following~\citet{dit_scalable_diffusion_Peebles_2023_ICCV}, we train DiT-XL/2 models for 400k steps and evaluate the generation FID (gFID) without classifier-free guidance~(CFG)~\citep{cfg_ho2021classifierfree} or, if specified, with a CFG of 1.375. 
We evaluate the throughput on a single H200 GPU, using the torch-compiled model for 1,000 iterations and a batch size of 128.
\looseness=-1

\mypar{Baselines}
We evaluate our model against the standard KL-VAE from \citet{sd1_Rombach_2022_CVPR} trained on ImageNet, and the  commonly-used stronger performing ft-EMA VAE~\citep{dit_scalable_diffusion_Peebles_2023_ICCV}, which we refer to as \sdvae{}.
We also include LiteVAE~\citep{litevae_NEURIPS2024_0735ab35} and VA-VAE~\citep{vavae_Yao_2025_CVPR},
and compare against existing diffusion decoders
\Dito{}~\citep{dito_chen2025diffusionautoencodersscalableimage} and \evae{}~\citep{evae_zhao2025epsilonvaedenoisingvisualdecoding}.
We evaluate the publicly-available weights of SD-VAE (f8c4) and the KL-VAE (f8c4, f16c16, f32c64), and reproduce \Dito{} training from their codebase.
For other baselines, we use available published results by lack of public code or weights.

\input{tables/sota_comp_reconstruction_vary_encoder}

\subsection{Image reconstruction and generation}
\label{subsec:exp_img_rec_gen}

\mypar{Fast image reconstruction}
We compare reconstruction quality of our \Ssdd{} models against other baselines in \Cref{tab:sota_comp_reconstruction_vary_encoder} using three encoder configurations. 
All methods use the same encoder architecture, and we use single-step distilled models for \Ssdd{}. 
Our method outperforms all existing  in terms of rFID, LPIPS and DreamSim, across all encoder configurations, using smaller models and a single sampling step.
We also benefit from a significant throughput increase compared to multi-step models, while maintaining high reconstruction quality.
In particular, for the {\bf f8c4} encoder configuration, our \Ssdd{}-S model \textbf{outperforms all single-step decoding methods both in speed and perceptual quality}, including single-step sampling from \evae{}.
Scaling the number of parameters to 345.9M, our \Ssdd{}-H obtains similar or better rFID compared to \evae{}-H, while benefiting from a $3\times$ increase in decoding speed.
In \Cref{fig:main_fig} we compare  deterministic and generative reconstruction methods in a speed-rFID plot, showing that \Ssdd{} is Pareto optimal across the board.

Switching to  \textbf{f16c16} and \textbf{f16c4} encoders, the standard KL-VAE decoders are deeper and narrower, reducing their parameter count with an associated drop in quality.
Since our models keep similar size and speed across the different encoder configurations,
\Ssdd{} consistently outperforms existing baselines with a higher compression ratio (\textbf{f16c4}), while maintaining similarly high throughput.
Using these deeper encoders, the gap of performance between deterministic and diffusion decoders widens. 
This reveals the impact of generative-focused reconstruction for heavily compressed representations.
We also observe an increased gap in sample quality between smaller and larger models, associated with their respective modeling capacity.
Larger models bring larger quality gains, as the conditional image distribution modeling task complexity increases.
Small models (\Ssdd{}-S) lack those generation capabilities, but achieve fast, high-quality reconstruction at lower compression rates.
\Ssdd{}-M strikes a balance by achieving good reconstruction quality at all scales with a parameter count similar to standard KL-VAE models.

While \Ssdd{} improves over prior state-of-the-art autoencoders in terms of rFID, LPIPS and DreamSim,  it yields somewhat worse results in terms of low-level distortion metrics (SSIM, PSNR).
We believe, however, that these metrics are not representative for the perceived reconstruction quality --- as illustrated in the  qualitative examples in \Cref{fig:main_fig}, \Cref{fig:qual_f8c4}, \Cref{fig:qual_f32c64} and \Cref{fig:qual_tradeoff} --- and should be de-emphasized for evaluation of generation-oriented decoders.

\input{tables/reconstruction}

\mypar{Scaling across model and image size}
We scale the input resolution from $128$ to $512$ and model size from \textbf{S} to \textbf{H} in \Cref{tab:reconstruction}, using the  f8c4 encoder configuration.
Our \Ssdd{} models use  single-step distilled  decoders fine-tuned at either $128\!\times\!128$ (evaluations at $128$), or $256\!\times\!256$ (evaluations at $256$ and $512$).
Overall, \Ssdd{} yields significantly better results than existing baselines, even with smaller models, in particular at 128 and 512 resolution.

\input{parts/figures/spatial_downsampling}

\input{tables/generation}

\mypar{Impact of spatial downsampling}
We conduct evaluations of reconstruction quality on both \Ssdd{} and KL-VAE across encoders with different spatial downsampling rates, but using the same total latent space size. 
In particular, we consider three settings where  $f^2 / c=16$.
In \Cref{fig:deep_enc} we show reconstruction quality as a function of the downsampling factor.
We also visualize reconstructions at the lowest and highest spatial downsampling factors in \Cref{fig:deep_enc}, and for more examples in \Cref{fig:qual_f8c4,fig:qual_f32c64}.
For  \Ssdd{} we observe  a modest decrease in quality when increasing $f$, whereas  KL-VAE suffers from a severe  degradation in performance for the $f=32$ setting.
This confirms that \Ssdd{} architecture functions well for varying levels of  spatial downsampling with limited  effect on resulting reconstructions.

\mypar{Application on image generation}
We conduct image generation experiments by training DiT-XL models~\citep{dit_scalable_diffusion_Peebles_2023_ICCV} on ImageNet $256\!\times\!256$.
In  \Cref{tab:generation} we report results in terms of rFID and generation speed.
We observe that the increased generative capabilities of the \Ssdd{} decoders results in higher sample quality across all setting compared to KL-VAE and \Dito{}, with and without CFG.
While our smallest model \Ssdd{}-S outperforms all baselines, larger decoders progressively increase the image quality at all encoder configurations. 
Additionally, we show that \Ssdd{} can significantly increase the throughput for similar generation quality.
Using a f16c4 encoder, the image is compressed into a $4\times$ smaller embedding compared to the f8c4 setting.
Using a larger decoder (\Ssdd{}-L) with DiT-XL/2, we can generate images $3.8\times$ faster than with the same DiT based on KL-VAE with an f8c4 encoder, with no loss in quality both with and without CFG.

\input{tables/ablation_from_dito}

\mypar{Ablation of our decoder design choices}
In \Cref{tab:ablation_from_dito}, we evaluate the impact of each of our design choices.
We start from a \Dito{}-S model as a baseline: we reproduce the \Dito{} architecture from~\citet{dito_chen2025diffusionautoencodersscalableimage},
and reduce the base channel number to $64$, yielding a 48.3M parameters (similar to \Ssdd{}-M) decoder with an f8c4 encoder, which we refer to as \Dito{}-S.
We train this model  on $128\!\times\!128$ images, using the same optimizer and parameters as \Ssdd{}.
We then progressively add components of our method.
Each one improves the generative capabilities of the model (rFID), with some increasing the image-to-image distortion.
In particular, our improved architecture and its regularization with the REPA loss bring the most improvements of 1.16 and 0.44 rFID respectively. 
Using a KL-encoder instead of the layer norm of~\citet{dito_chen2025diffusionautoencodersscalableimage} has a positive impact on performance, and allows us to use a standard encoder as commonly used to train generative models~\citep{sd3_pmlr-v235-esser24a,dit_scalable_diffusion_Peebles_2023_ICCV}.
It focuses our work around the \textit{decoder} component, without requiring specific properties from the encoder.
Additionally, the use of a shared encoder (\Cref{subsec:encoder}) and shared pre-training (stage 1 in \Cref{subsec:method_training}), which improve training efficiency, do not hurt performance.
Finally, only the distillation has a slight but negative impact on rFID, but brings the model from 8-steps to 1-step sampling.
To demonstrate that we can achieve optimal results with a GAN-free method, we also evaluate the impact of adding a GAN loss.
We follow the parametrization of the \textit{adversarial denoising trajectory matching} from~\citep{evae_zhao2025epsilonvaedenoisingvisualdecoding}, and use it in combination with the setup where we added the shared encoder, shown three lines above and indicated by $\S$.
We do not observe any significant impact on rFID and other metrics.
A more detailed analysis of the impact of the GAN is included in \Cref{app:qualitative_results}.
We further isolate the contribution of each ablation component into essential, standard-recipe and pipeline-only groups in \Cref{app:component_isolation}, and report additional ablations on the GAN loss applied to the distilled single-step model and on L2-only distillation in \Cref{app:extended_ablations} (\Cref{tab:app_extended_ablations}). \Cref{app:meanflow} additionally compares our two-stage pipeline to single-stage MeanFlow training.

\mypar{Evaluation on COCO images}
To assess the generalization of our model to other sets of natural images, we follow~\cite{evae_zhao2025epsilonvaedenoisingvisualdecoding} by evaluating \Ssdd{} and comparable baselines on the 5k images from the COCO-2017 evaluation set~\citep{coco}. Results are reported in \Cref{tab:rec_coco}. We observe that \Ssdd{} generalizes better than existing baselines. 
In particular, \Ssdd{}-B yields better reconstruction than \evae{} (H), while it was only surpassed by \Ssdd{}-L and \Ssdd{}-XL on ImageNet in \Cref{tab:reconstruction}. This shows that, despite a relatively small size and high decoding speed, our models provide high reconstruction fidelity at various model scales and data.

\input{tables/rec_coco}

%% file: tables/sota_comp_reconstruction_vary_encoder.tex
\begin{table}[t]
\caption{
    {\bf Comparison with state-of-the-art models on ImageNet $\bf 256\!\times\!256$.} 
    Using  a downsampling factor $f$ of 8 or 16, and $c\!=\!4$ or $c\!=\!16$  latent channels. 
    For each metric, we highlight the \textbf{first} and \underline{second} best results.
    \textit{\#D: decoder parameter count.}
    \textit{N: number of decoding steps.}
    \textit{$\upsymb{2}$: numbers reported from original papers.}
    \textit{$\upsymb{3}$: SD-VAE is the ft-EMA fine-tuned variant of KL-VAE on a larger dataset~\citep{dit_scalable_diffusion_Peebles_2023_ICCV}.}
}
\label{tab:sota_comp_reconstruction_vary_encoder}
\begin{center}
\scriptsize

\resizebox{\textwidth}{!}{
\begin{tabular}{l|lr|rr|ccccc}
\toprule

& \textbf{Method}
& \textbf{\#D}
& \textbf{$\mathbf{N}$}
& \textbf{Ims./sec.}
& \textbf{rFID\odec}
& \textbf{LPIPS\odec}
& \textbf{DreamSim\odec}
& \textbf{PSNR\oinc}
& \textbf{SSIM\oinc} \\

\midrule

\multirow{13}{*}{\rotatebox{90}{\textbf{f8c4}}}
 & \sdvae{}$^{\upsymb{3}}$ & 47.2M & 1 & \underline{707} & 0.69 & 0.061 & 0.042 & 25.52 & 0.78 \\
 & KL-VAE~\citep{sd1_Rombach_2022_CVPR} & 47.2M & 1 & 705 & 0.87 & 0.065 & 0.046 & 24.11 & 0.75 \\
 & LiteVAE~\citep{sd1_Rombach_2022_CVPR}$^{\upsymb{2}}$ & 53.3M & 1 & \underline{707} & 0.87 & - & - & 26.02 & 0.74 \\
 & \evae{} (B, 1-step)~\citep{evae_zhao2025epsilonvaedenoisingvisualdecoding}$^{\upsymb{2}}$ & 20.6M & 1 & 526 & 0.57 & - & - & - & - \\
 & \evae{} (M)~\citep{evae_zhao2025epsilonvaedenoisingvisualdecoding}$^{\upsymb{2}}$ & 49.3M & 3 & 134 & 0.47 & - & - & \underline{27.65} & \underline{0.84} \\
 & \evae{} (H)~\citep{evae_zhao2025epsilonvaedenoisingvisualdecoding}$^{\upsymb{2}}$ & 355.6M & 3 & 33 & 0.38 & - & - & \textbf{29.49} & \textbf{0.85} \\
 & \Dito{}-XL-LPIPS~\citep{dito_chen2025diffusionautoencodersscalableimage} & 592.2M & 50 & 1 & 1.06 & 0.077 & 0.055 & 23.53 & 0.72 \\
 & Consistency decoder~\citep{dalle3}$^{\upsymb{2}}$ & 625.1M & 2 & 23 & 0.80 & 0.065 & 0.044 & 23.71 & 0.73 \\
 & \ssddone{-S} & 13.4M & 1 & \textbf{1027} & 0.46 & 0.060 & 0.039 & 24.08 & 0.74 \\
 & \ssddone{-B} & 20.2M & 1 & 689 & 0.42 & 0.058 & 0.036 & 24.18 & 0.75 \\
 & \ssddone{-M} & 48.0M & 1 & 357 & 0.39 & 0.055 & 0.034 & 24.38 & 0.75 \\
 & \ssddone{-L} & 85.2M & 1 & 302 & \underline{0.36} & \underline{0.053} & \underline{0.032} & 24.49 & 0.76 \\
 & \ssddone{-XL} & 153.8M & 1 & 215 & \textbf{0.35} & \textbf{0.052} & \textbf{0.031} & 24.60 & 0.76 \\

\midrule

\multirow{8}{*}{\rotatebox{90}{\textbf{f16c16}}}
 & KL-VAE~\citep{sd1_Rombach_2022_CVPR} & 36.4M & 1 & \textbf{1196} & 0.82 & 0.066 & 0.048 & 23.88 & \underline{0.74} \\
 & VA-VAE~\citep{vavae_Yao_2025_CVPR}$^{\upsymb{2}}$ & 39.5M & 1 & \underline{1178} & 0.55 & 0.132 & - & \textbf{26.10} & 0.72 \\
 & FlowMo (continuous)~\citep{flow_to_the_mode_sargent2025flowmodemodeseekingdiffusion}$^{\upsymb{2}}$ & - & 25 & - & 0.65 & 0.054 & - & 26.61 & 0.791 \\
 & \ssddone{-S} & 13.4M & 1 & 985 & 0.45 & 0.061 & 0.038 & 23.75 & 0.73 \\
 & \ssddone{-B} & 20.2M & 1 & 686 & 0.41 & 0.059 & 0.035 & 23.86 & 0.73 \\
 & \ssddone{-M} & 48.0M & 1 & 368 & \underline{0.40} & {0.056} & \underline{0.033} & 24.08 & \underline{0.74} \\
 & \ssddone{-L} & 85.2M & 1 & 315 & \textbf{0.34} & \underline{0.053} & \textbf{0.030} & 24.20 & \underline{0.74} \\
 & \ssddone{-XL} & 153.8M & 1 & 213 & 0.36 & \textbf{0.052} & \textbf{0.030} & \underline{24.31} & \textbf{0.75} \\

\midrule

\multirow{8}{*}{\rotatebox{90}{\textbf{f16c4}}}
 & KL-VAE~\citep{sd1_Rombach_2022_CVPR} & 36.4M & 1 & \textbf{1176} & 2.93 & - & - & 20.57 & 0.66 \\
 & \evae{} (M)~\citep{evae_zhao2025epsilonvaedenoisingvisualdecoding}$^{\upsymb{2}}$ & 49.3M & 3 & 134 & 1.91 & - & - & \underline{21.27} & \underline{0.69} \\
 & \evae{} (H)~\citep{evae_zhao2025epsilonvaedenoisingvisualdecoding}$^{\upsymb{2}}$ & 355.6M & 3 & 33 & 1.35 & - & - & \textbf{22.60} & \textbf{0.71} \\
 & \ssddone{-S} & 13.4M & 1 & \underline{951} & 2.29 & 0.186 & 0.121 & 15.59 & 0.45 \\
 & \ssddone{-B} & 20.2M & 1 & 663 & 1.77 & 0.177 & 0.111 & 16.06 & 0.46 \\
 & \ssddone{-M} & 48.0M & 1 & 360 & {1.23} & 0.161 & 0.096 & 17.13 & 0.49 \\
 & \ssddone{-L} & 85.2M & 1 & 305 & \underline{0.96} & \underline{0.155} & \underline{0.089} & 17.26 & 0.50 \\
 & \ssddone{-XL} & 153.8M & 1 & 216 & \textbf{0.89} & \textbf{0.148} & \textbf{0.083} & 17.72 & 0.51

 \\ \bottomrule
\end{tabular}
}

\end{center}
\end{table}

%% file: tables/reconstruction.tex
\begin{table}[t]
\caption{
    \textbf{Scaling of models and image resolution on ImageNet.}
    All models use an f8c4 encoder.
    \Ssdd{} instances use the same shared encoder and are distilled into single-step decoders.
    Decoder with similar parameters count are shown highlighted with the same color.
    Evaluations at $512\times 512$ or higher resolution are conducted using the same model as for $256\times 256$.
}

\label{tab:reconstruction}
\begin{center}
\scriptsize

\begin{tabular}{lr|cc|cc|cc|cc}
\toprule

\multirow{2}{*}{\textbf{Method}}
    & \multirow{2}{*}{\textbf{\#D}}
    & \multicolumn{2}{c|}{\textbf{$128\!\times\!128$}}
    & \multicolumn{2}{c|}{\textbf{$256\!\times\!256$}}
    & \multicolumn{2}{c}{\textbf{$512\!\times\!512$}}
    & \multicolumn{2}{c}{\textbf{$1024\!\times\!1024$}}
    \\

    &
    & \textbf{rFID\odec}
    & \textbf{LPIPS\odec}
    & \textbf{rFID\odec}
    & \textbf{LPIPS\odec}
    & \textbf{rFID\odec}
    & \textbf{LPIPS\odec}
    & \textbf{rFID\odec}
    & \textbf{LPIPS\odec}
    \\

\midrule

\rowM KL-VAE & 47.2M & - & - & 0.87 & 0.065 & 0.29 & 0.064 & 0.20 & 0.059 \\
\rowB \evae{} (B) & 20.6M & 1.94 & - & 0.52 & - & 0.61 & - & - & - \\
\rowM \evae{} (M) & 49.3M & 1.58 & - & 0.47 & - & 0.53 & - & - & - \\
\rowL \evae{} (L) & 89.0M & 1.47 & - & 0.45 & - & 0.41 & - & - & - \\
\rowXL \evae{} (XL) & 140.6M & 1.34 & - & 0.43 & - & 0.39 & - & - & - \\
\rowH \evae{} (H) & 355.6M & 1 & - & 0.38 & \textbf{-} & 0.35 & - & - & - \\
\rowXH \Dito{}-XL + LPIPS & 592.2M & - & - & 1.06 & 0.077 & 0.28 & 0.080 & 0.15 & 0.086 \\
\rowXH Consistency decoder & 625.1M & - & - & 0.80 & 0.065 & 0.30 & 0.066 & 0.15 & 0.061 \\

\midrule

\rowS\ssddone{-S} & 13.4M & 1.89 & 0.061 & 0.46 & 0.060 & 0.28 & 0.062 & 0.16 & 0.064 \\
\rowB\ssddone{-B} & 20.2M & 1.48 & 0.059 & 0.42 & 0.058 & 0.22 & 0.060 & 0.13 & 0.058 \\
\rowM\ssddone{-M} & 48.0M & 1.04 & 0.056 & 0.39 & 0.055 & \underline{0.20} & 0.057 & \underline{0.12} & 0.055 \\
\rowL\ssddone{-L} & 85.2M & \underline{0.88} & \underline{0.054} & \underline{0.36} & \underline{0.053} & \textbf{0.19} & \underline{0.055} & \textbf{0.11} & \underline{0.053} \\
\rowXL\ssddone{-XL} & 153.8M & \textbf{0.81} & \textbf{0.052} & \textbf{0.35} & \textbf{0.052} & \underline{0.20} & \textbf{0.053} & \textbf{0.11} & \textbf{0.051}

 \\ \bottomrule
\end{tabular}

\end{center}
\end{table}

%% file: parts/figures/spatial_downsampling.tex
\begin{figure}[t!]
\centering
    \begin{subfigure}{.47\textwidth}
      \centering
        \includegraphics[width=\textwidth]{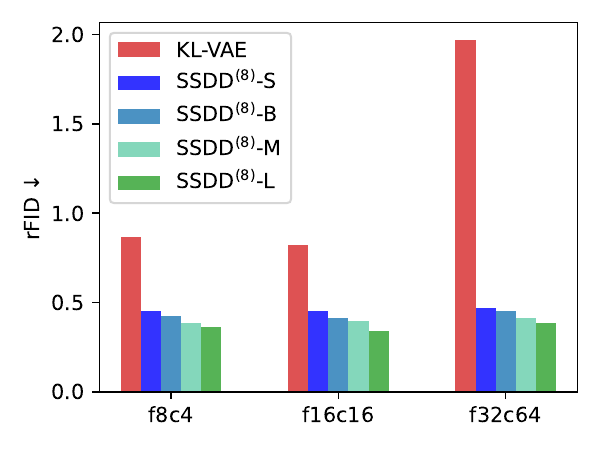}
        \label{fig:deep_enc_1}
        \tabfigendspace
    \end{subfigure}
    \begin{subfigure}{.5\textwidth}
      \centering
        \includegraphics[width=\textwidth]{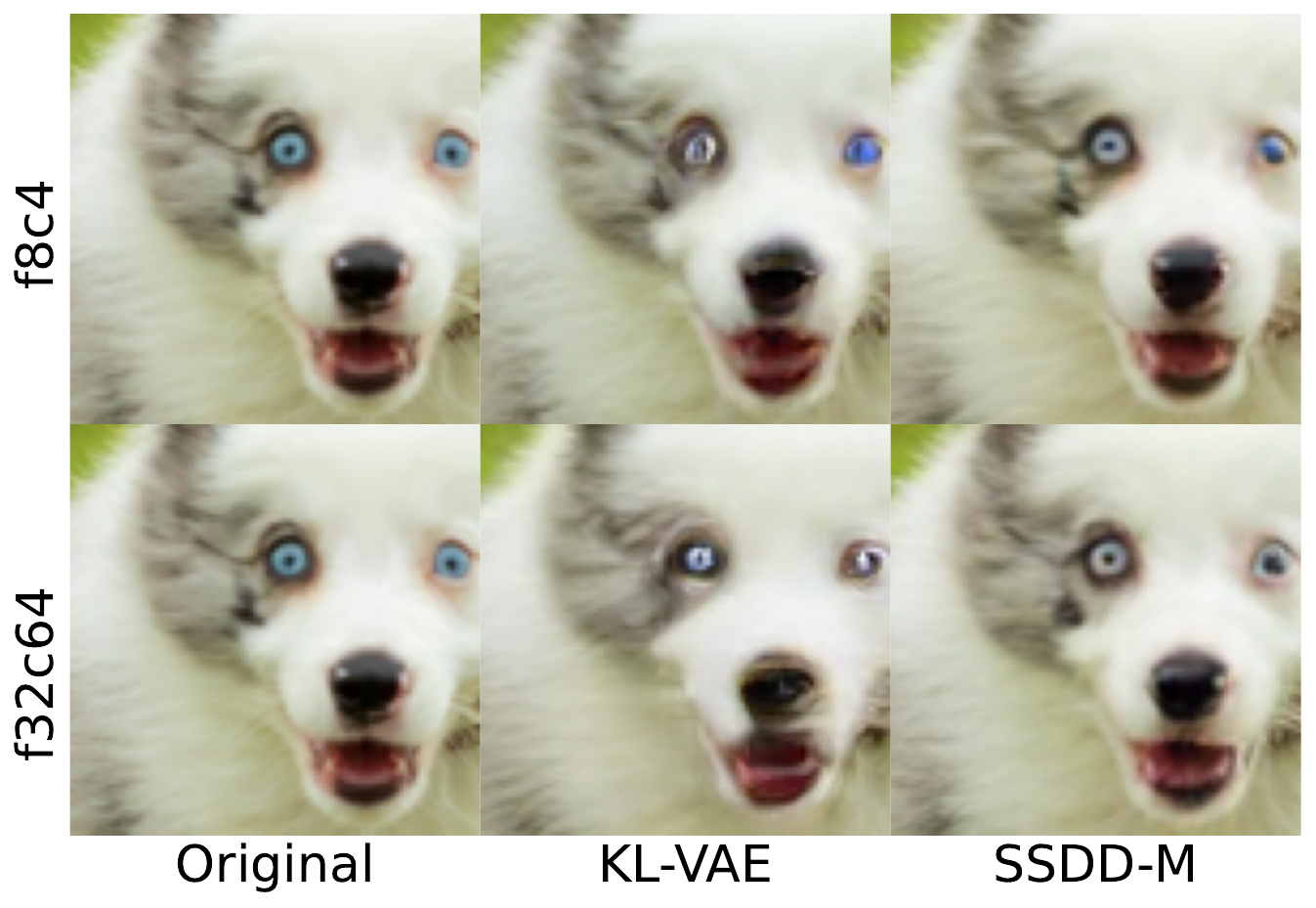}
        \label{fig:deep_enc_2}
        \tabfigendspace
    \end{subfigure}
    \caption{
        \textbf{Evolution of  rFID and qualitative comparison when increasing spatial downsampling.}
        Evaluated on ImageNet $256\!\times\!256$ with a constant compression ratio by adjusting $c$. Images patches are high-frequency features extracted from a larger $256\times 256$ image.
    }
    \label{fig:deep_enc}
\end{figure}

%% file: tables/generation.tex
\begin{table}[t!]
\caption{
    \textbf{Evaluation of autoencoders on image generation on ImageNet $256\times 256$.}
    Generation speed in images/second includes  sampling the DiT-XL/2 model as well as pixel decoding.
}
\label{tab:generation}
\begin{center}
\scriptsize

\begin{tabular}{Hc|l|rr|rr}
\toprule

& &  \multirow{2}{*}{\textbf{Autoencoder}} &  \multicolumn{2}{c}{\textbf{No CFG}} &  \multicolumn{2}{c}{\textbf{With CFG}} \\
& & &
    \textbf{gFID\odec} &
    \textbf{Images/s} &
    \textbf{gFID\odec} &
    \textbf{Images/s} \\

\midrule

    & \multirow{7}{*}{\rotatebox{90}{\textbf{f8c4}}}  
    & KL-VAE~\cite{sd1_Rombach_2022_CVPR}  & 19.71 & \underline{7.05} & 7.63 & \underline{3.54}\tikzmark{a1}  \\
 &  & \Dito{}-XL-LPIPS~\cite{dito_chen2025diffusionautoencodersscalableimage} & 22.47 & 0.87 & 10.83 & 0.77 \\
 &  & \ssddone{-S} & 17.77 & \bf 7.07 & 8.02 & \bf 3.55 \\
 &  & \ssddone{-B} & 17.46 & \underline{7.05} & 7.85 & \underline{3.54} \\
 &  & \ssddone{-M} & 17.03 & 6.98 & 7.63 & 3.52 \\
 &  & \ssddone{-L} & \underline{16.86} & 6.96 & \underline{7.48} & 3.52 \\
 &  & \ssddone{-XL} & \bf 16.79 & 6.89 & \bf 7.46 & 3.50 \\

\cmidrule{2-7}
 & \multirow{5}{*}{\rotatebox{90}{\textbf{f16c4}}}  
    & \ssddone{-S} & 20.28 & \bf 27.55 & 9.83 & \bf 13.98 \\
 &  & \ssddone{-B} & 18.89 & \underline{27.20} & 8.98 & \underline{13.89} \\
 &  & \ssddone{-M} & 17.24 & 26.29 & 7.99 & 13.65 \\
 &  & \ssddone{-L} & \underline{16.27} & 25.95 & \underline{7.34} & 13.55\tikzmark{t1} \\
 &  & \ssddone{-XL} & \bf{16.08} & 25.04 & \bf 7.12 & 13.30 \\

\cmidrule{2-7}
    & \multirow{6}{*}{\rotatebox{90}{\textbf{f16c16}}}

    & KL-VAE~\cite{sd1_Rombach_2022_CVPR} & 31.96 & \bf 27.79 & 26.42 & \bf 14.04 \\
 &  & \ssddone{-S} & 25.84 & \underline{27.57} & 20.51 & \underline{13.98} \\
 &  & \ssddone{-B} & 25.59 & 27.24 & 20.28 & 13.90 \\
 &  & \ssddone{-M} & 25.20 & 26.34 & 19.91 & 13.66 \\
 &  & \ssddone{-L} & \underline{25.03} & 26.02 & \underline{19.74} & 13.57 \\
 &  & \ssddone{-XL} & \bf 25.01 & 25.07 & \bf 19.68 & 13.31

 \\ \bottomrule
\end{tabular}

\begin{tikzpicture}[overlay, remember picture, shorten >=.5pt, shorten <=.5pt, transform canvas={yshift=.25\baselineskip,xshift=.25em}]
\draw [-{Stealth[length=2.4mm, width=1.6mm]}, red] ({pic cs:a1}) [bend left] to node [below right] (t1) {\hspace{2pt}\scriptsize \textbf{3.8$\times$}} ({pic cs:t1});
\end{tikzpicture}

\end{center}
\end{table}

%% file: tables/ablation_from_dito.tex
\newcommand{\abgood}[1]{\textit{\color{OliveGreen}#1}}
\newcommand{\abbaad}[1]{\textit{\color{Red}#1}}
\newcommand{\abneut}[1]{\textit{\color{Gray}±#1}}

\begin{table}[t]
\caption{
    {\bf Ablation of design choices starting from \Dito{} as baseline.}
    We indicate the number $N$ of sampling steps used for each model, directly affecting decoding speed.
}
\label{tab:ablation_from_dito}
\begin{center}
\scriptsize

\begin{tabular}{l|rlrlrl|c}
\toprule

\textbf{Ablation} & 
\multicolumn{2}{l}{\textbf{rFID\odec} } & 
\multicolumn{2}{l}{\textbf{PSNR\oinc}}& 
\multicolumn{2}{l|}{\textbf{DreamSim\odec}}&
$N$\\

\midrule

\Dito{}-S-LPIPS \textit{(\#D=48.3M)}  & 3.17  &  & 23.10 &  & 0.107 &  & 24 \\
$\vert$ + \Ssdd-M decoder \textit{(\#D=48.0M)}   & 2.01  & \abgood{-1.16}  & 23.38 & \abgood{+0.28}  & 0.060 & \abgood{-0.048} & 24 \\
$\vert$ + REPA loss                & 1.58  & \abgood{-0.44}  & 22.82 & \abbaad{-0.56} & 0.104 & \abbaad{+0.045}  & 24 \\
$\vert$ + replace z-norm by KL     & 1.32  & \abgood{-0.26}  & 23.02 & \abgood{+0.20}  & 0.098 & \abgood{-0.006} & 24 \\
$\vert$ + logit-normal sampling    & 1.30  & \abgood{-0.02}  & 23.13 & \abgood{+0.11}  & 0.101 & \abbaad{+0.003}  & 16 \\
$\vert$ + $t$-spacing sampler        & 1.25  & \abgood{-0.05}  & 23.43 & \abgood{+0.30}  & 0.097 & \abgood{-0.004} & 8  \\
$\vert$ + EMA                & 1.17  & \abgood{-0.08}  & 23.38 & \abbaad{-0.05} & 0.097 & \abneut{0.000} & 8  \\
$\vert$ + shared encoder ($\S$)    & 1.07  & \abgood{-0.10}  & 23.58 & \abgood{+0.20}  & 0.090 & \abbaad{+0.030}  & 8  \\
$\vert$ + shared pre-training 
\textbf{(\ssddmulti$^{(8)}$-M)}  & \textbf{0.97}  & \abgood{-0.10}& 23.04 & \abbaad{-0.54}& 0.087 & \abgood{-0.003} & 8 \\
$\vert$ + distillation \textbf{(\Ssdd-M)} & 1.04  & \abbaad{+0.07}& 23.28   & \abgood{+0.24}& \textbf{0.084} & \abgood{-0.002} & 1 \\
\midrule
\textit{\textbf{($\S$)} + GAN loss}     & \textit{1.07} & \abneut{0.00} & \textit{23.58} & \abneut{0.00} & \textit{0.089} & \abgood{-0.001} & 8

 \\ \bottomrule
\end{tabular}

\end{center}
\end{table}

%% file: tables/rec_coco.tex
\begin{table}[t]
\caption{
    \textbf{Generalization of auto-encoders to new datasets.}
    All models use an f8c4 encoder and are trained at $256\times 256$ resolution on ImageNet, except for the consistency decoder. Evaluation  conducted on the COCO-2017 5K validation set.
}
\label{tab:rec_coco}
\begin{center}
\scriptsize

\begin{tabular}{lr|cc|cc}
\toprule

\multirow{2}{*}{\textbf{Method}}
    & \multirow{2}{*}{\textbf{\#D}}
    & \multicolumn{2}{c|}{\textbf{COCO $256\!\times\!256$}}
    & \multicolumn{2}{c}{\textbf{COCO $512\!\times\!512$}}
    \\

    &
    & \textbf{rFID\odec}
    & \textbf{LPIPS\odec}
    & \textbf{rFID\odec}
    & \textbf{LPIPS\odec}
    \\

\midrule

\rowM KL-VAE~\citep{sd1_Rombach_2022_CVPR} & 47.2M & 4.65 & 0.063 & 2.68 & 0.064 \\
\rowM \evae{} (M)~\citep{evae_zhao2025epsilonvaedenoisingvisualdecoding} & 49.3M & 3.98 & - & - & - \\
\rowH \evae{} (H)~\citep{evae_zhao2025epsilonvaedenoisingvisualdecoding} & 355.6M & 3.65 & - & - & - \\
\rowXH \Dito{}-XL + LPIPS~\citep{dito_chen2025diffusionautoencodersscalableimage} & 592.2M & 5.95 & 0.076 & 2.93 & 0.079 \\
\rowXH Consistency decoder~\citep{dalle3} & 625.1M & 4.53 & 0.063 & 2.65 & 0.065 \\
\rowS\ssddone{-S} & 13.4M & 3.77 & 0.059 & 2.51 & 0.062 \\
\rowB\ssddone{-B} & 20.2M & 3.62 & 0.057 & 2.34 & 0.059 \\
\rowM\ssddone{-M} & 48.0M & 3.41 & 0.054 & 2.24 & 0.056 \\
\rowL\ssddone{-L} & 85.2M & \underline{3.29} & \underline{0.052} & \underline{2.15} & \underline{0.054} \\
\rowXL\ssddone{-XL} & 153.8M & \textbf{3.25} & \textbf{0.050} & \textbf{2.13} & \textbf{0.053}

 \\ \bottomrule
\end{tabular}

\end{center}
\end{table}

%% file: parts/concl.tex
\section{Conclusion}

We introduce \Ssdd{}, a diffusion autoencoder that leverages flow-matching, perceptual alignment and REPA regularization to reach state-of-the-art reconstruction and high generation quality without relying on adversarial training.  
Through a lightweight distillation strategy, we show that multi-step diffusion behavior can be compressed into a single-step decoder, yielding up to $10\times$ faster decoding than 
\mbox{\evae{} (H)} at similar rFID. On ImageNet, \Ssdd{} consistently outperforms KL-VAE and recent diffusion decoders across reconstruction (rFID, LPIPS, DreamSim) and downstream generation (gFID with DiT), with particularly strong gains under high-compression settings.
These results establish \Ssdd{}  as the first GAN-free single-step diffusion decoder to reach state-of-the-art continuous tokenization, combining speed, stability, and generative fidelity, and providing a scalable foundation for future large-scale generative modeling.

\section*{Acknowledgement}
This work has been supported by chair VISA DEEP (ANR-20-CHIA-0022) and Cluster PostGenAI@Paris (ANR-23-IACL-0007, FRANCE 2030).

%% file: parts/supmat.tex
\appendix
\onecolumn
\clearpage

\setcounter{figure}{0} 
\renewcommand\thefigure{S\arabic{figure}} 
\setcounter{table}{0}
\renewcommand{\thetable}{S\arabic{table}}

\theoremstyle{plain}
\newtheorem{theorem}{Theorem}[section]
\newtheorem{lemma}[theorem]{Lemma}
\newtheorem{proposition}[theorem]{Proposition}
\newtheorem{corollary}[theorem]{Corollary}
\theoremstyle{definition}
\newtheorem{definition}[theorem]{Definition}
\newtheorem{example}[theorem]{Example}
\theoremstyle{remark}
\newtheorem{remark}[theorem]{Remark}
\newtheorem{note}[theorem]{Note}
\newtheorem{statement}{Statement}[section]

\section{Theoretical motivations}
\label{app:theory}

As discussed in~\Cref{sec:intro}, autoencoders have to navigate a  fundamental trade-off between image-to-image \textbf{distortion} $\Delta(x,\hat{x})$ that can be minimized by a deterministic decoder,
and \textbf{distribution shift} $d(P_x,P_{\hat{x}})$, best optimized using a generative decoder.
We show here an illustration of this principle on a synthetic example, followed by theoretical justifications of the previous statement.
For a more formal analysis of this trade-off, we refer to \citet{Blau_2018_CVPR}.

\mypar{Motivation: synthetic example}
Let us illustrate this trade-off with a one dimensional synthetic data example.
Let $P_x=\mathcal{U}(-2,2)$ be the source distribution that takes the form of a uniform distribution over $[-2;2]$, and  $E_d(x)=\fnsign(x) \in \{-1,1\}$ the encoder that maps source samples to their sign.
For the distortion metric $\Delta(x,\hat{x})=(x-\hat{x})^2$ the optimal decoder is $D^s(z)=z$.
We obtain this result by minimization of $\int_0^2(D^s(1)-x)^2dx$ and $\int_{-2}^0(D^s(-1)-x)^2dx$, which also holds if $D^s(z)$ is a random variable.
Using the Kullback–Leibler divergence as the distribution shift metric $d_{KL}(P_x\Vert P_{\hat{x}})$,
the non-deterministic decoder $D^g(z)\sim\mathcal{U}(z-1,z+1)$ achieves an optimal score.
Indeed, using Gibbs' inequality, $d_{KL}(P_x\Vert P_{\hat{x}})$ is $0$ if and only if $P_x = P_{\hat{x}}$, which is in particular true by using $D^g$, and can only be achieved by a non-deterministic decoder when using $E$ as the encoder.
Additionally, computing the distortion and distribution shift for $D^s$ and $D^g$ gives that $D^s$ has a strictly lower distortion and strictly higher distribution shift than $D^g$.
This synthetic example illustrates how generative decoders can help navigate the trade-off between both metric groups.

\mypar{Distortion-distribution shift trade-off: supporting claims}
We also introduce the following simple claims to support the intuition behind generative decoding by giving theoretical results for boundary cases:
\begin{itemize}
    \item \textbf{There is always a deterministic decoder minimizing a given distortion metric.}
        We assume $x$ and the decoder outputs are contained in a closed set $X$.
        If a non-deterministic decoder $D^g(z)$ minimizes distortion, for any $z$, we have:
        \begin{align}
            &\int_{x\in E^{-1}(z)}\int_{y\in X}\Delta(x,y)P(D^g(z)=y)dydx \\
            &\geq \int_{y\in X} \min_{\hat{y}} \left[ \int_{x\in E^{-1}(z)}\Delta(x,\hat{y})\right ]P(D^g(z)=y)dydx \\
            &=\min_{\hat{y}} \left[ \int_{x\in E^{-1}(z)}\Delta(x,\hat{y})\right ]
        \end{align}
        which yields a deterministic decoder that outputs the minimizer $\hat{y}$ for any $z$.
    \item \textbf{There is always a non-deterministic decoder minimizing a given distribution shift metric}.
        We assume a metric such that $P=Q\implies d(P,Q)$ is minimal. Then we use the generative decoder $D^g(z)=u$ with $u\sim P_x$ a random variable, independent of $z$.
    \item \textbf{With a lossy encoder, only a non-deterministic decoder will perfectly model $P_x$}.
        This result comes directly from applying information theory.
        We define, for a random variable $x\in X$, a lossy encoder as $E:x\mapsto z\in Z$ such that $H(E(x)) < H(x)$, with $H$ measuring the entropy.
        Given a deterministic decoder $D^s$ and $\hat{x}=D^s(z)$, we have $H(\hat{x})\leq H(z)$ (deterministic functions cannot increase entropy), yielding $H(\hat{x})< H(x)$.
        As a result, $P_{\hat{x}}\neq P_x$.
\end{itemize}
These claims do not constitute a general proof that, for images, currently existing non-deterministic decoders offer a strictly better trade-off along the distortion-distribution shift curve.
But instead they offer intuitions and motivations for use of generative decoders.

\section{Implementation details}
\label{app:implementation_details}

\mypar{Model configurations}
For the encoder $E$ we follow the standard convolutional architecture from~\citet{sd1_Rombach_2022_CVPR}.
Our \Ssdd{} decoder architecture is detailed in~\Cref{subsec:method_architecture}.
We display in \Cref{tab:model_configs} the configuration for each of its blocks: the number of channels at each level ($1\!\times\!1$, $2\!\times\!2$, $4\!\times\!4$ and $8\!\times\!8$ patches) is the base \textbf{Channels} times the \textbf{Depth multiplier}.
The middle transformer operates with the same number of channels as the deepest ResNet, and we indicate the number of transformer blocks.
For each  decoder configuration  we also report the number of parameters as \#D.

\input{tables/model_configs}

\mypar{Training \& evaluation}
We train  our models  on ImageNet-1k~\citep{deng09cvpr} using the RAdamW schedule-free optimizer~\citep{schedule_free_the_road_less_scheduled}, a weight decay of $0.001$ and no scheduler.
We use the following loss coefficients: $\lambda_{\LPIPSFN}=0.5$, $\lambda_{\REPAFN}=0.25$, $\lambda_{KL}=10^{-6}$.
We maintain an exponential moving average of the weights with decay rate $0.999$, starting from 50k iterations.
During the first shared pre-training stage, we train on $128\!\times\!128$ crops with a constant learning rate of $3\cdot10^{-4}$ for 1M iterations.
When jointly training the encoder, we use a learning rate of $10^{-4}$ for increased stability.
During the second stage, we fine-tune the weights for 500k iterations, additionally decreasing the learning rate to $10^{-4}$ at the $256\!\times\!256$ resolution.
We use random resizing of the image in the range 128 to 256 during the first stage training, and resizing at the target resolution during the second stage,
to which we add random cropping at training resolution ($128$ or $256$) and horizontal flip augmentations.
All training resizing operations use Lanczos interpolation for downsampling, while evaluation uses bilinear interpolation to match standard evaluation settings~\citep{evae_zhao2025epsilonvaedenoisingvisualdecoding,dito_chen2025diffusionautoencodersscalableimage}.
For distillation, we use a 7-steps teacher model, to maintain a balance between reconstruction and generative capabilities (see \Cref{app:sampling_distill}).
To ensure sharp details, we distill for 50k iterations.

\section{Sampling and distillation analysis}
\label{app:sampling_distill}

\begin{figure*}[b!]
\centering
    \begin{subfigure}{.33\textwidth}
      \centering
        \includegraphics[width=\textwidth,trim={0pt 0pt 10pt 10pt},clip]{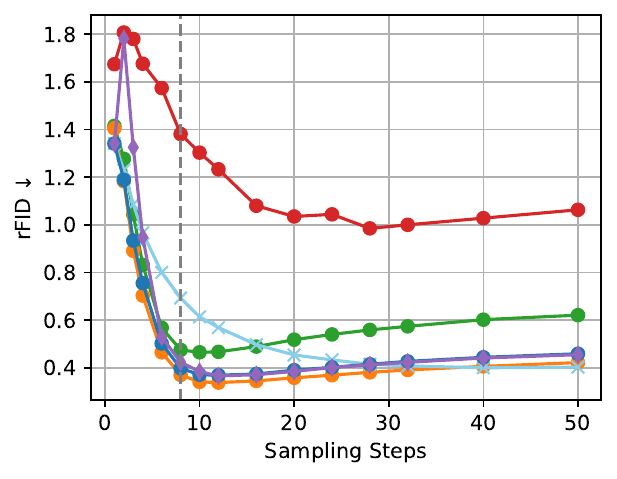}
        \label{fig:sampling_1}
        \tabfigendspace
    \end{subfigure}
    \begin{subfigure}{.33\textwidth}
      \centering
        \includegraphics[width=\textwidth,trim={0pt 0pt 10pt 10pt},clip]{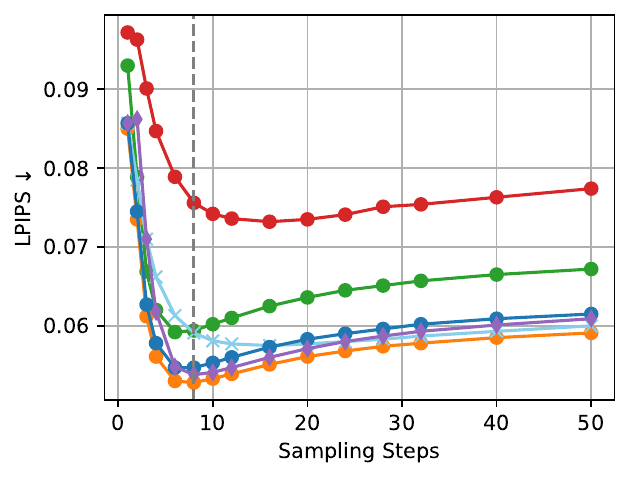}
        \label{fig:sampling_2}
        \tabfigendspace
    \end{subfigure}
    \begin{subfigure}{.33\textwidth}
      \centering
        \includegraphics[width=\textwidth,trim={0pt 0pt 10pt 10pt},clip]{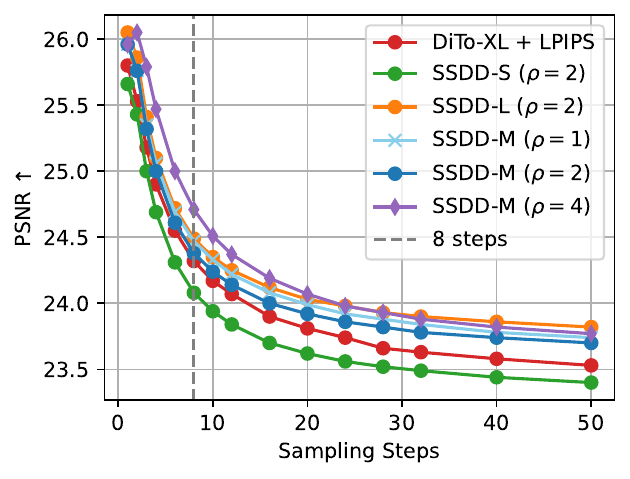}
        \label{fig:sampling_3}
        \tabfigendspace
    \end{subfigure}
    \caption{
        \textbf{Evolution of reconstruction metrics depending on the number of sampling steps $N$.}
        Evaluated on ImageNet \res{256}.
    }
    \label{fig:sampling}
\end{figure*}

\begin{figure*}[b!]
\centering
    \begin{subfigure}{.5\textwidth}
      \centering
        \includegraphics[width=\textwidth]{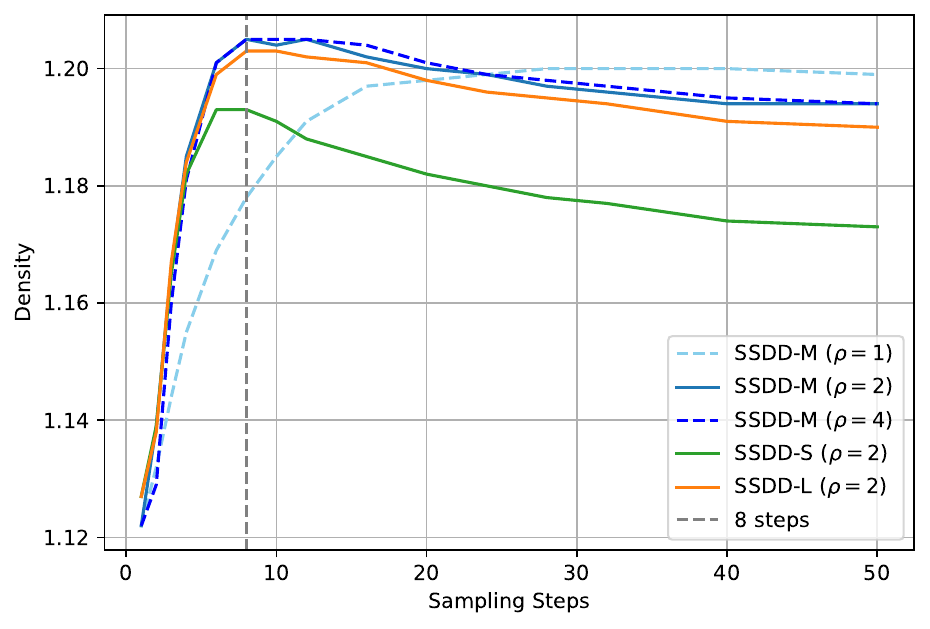}
        \label{fig:density}
        \tabfigendspace
    \end{subfigure}
    \begin{subfigure}{.5\textwidth}
      \centering
        \includegraphics[width=\textwidth]{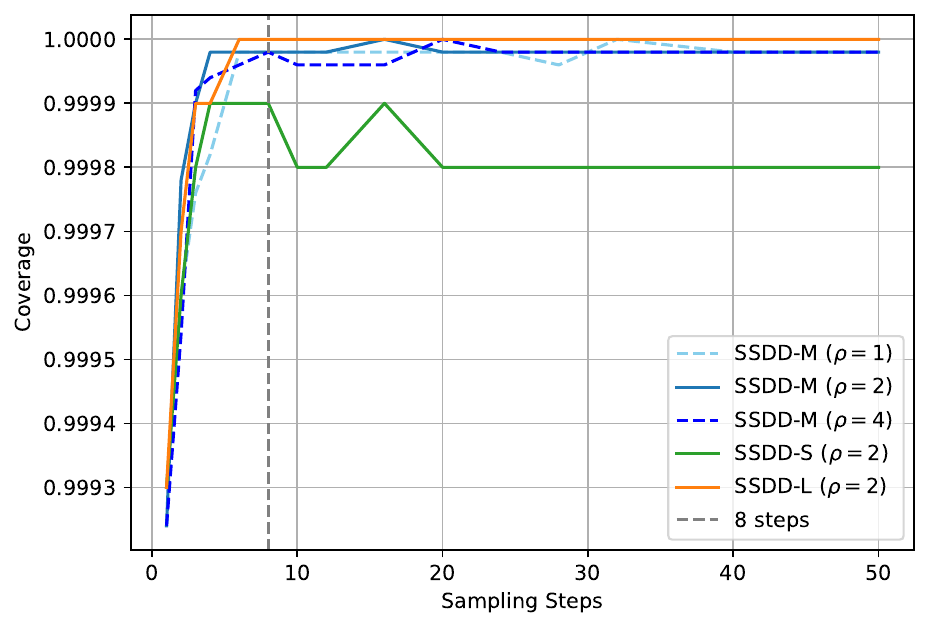}
        \label{fig:coverage}
        \tabfigendspace
    \end{subfigure}
    \caption{
        \textbf{Effect of sampling on Density and coverage.}
    }
    \label{fig:density_coverage}
\end{figure*}

\mypar{Impact of sampling steps on reconstruction}
Quality of samples from diffusion models usually improves with a higher number of sampling steps.
But as noted by~\citet{evae_zhao2025epsilonvaedenoisingvisualdecoding}, diffusion decoders have an \textit{optimal} number of sampling steps, after which reconstruction quality degrades.
In \Cref{fig:sampling} we display this effect on \Ssdd{} and \Dito{}~\citep{dito_chen2025diffusionautoencodersscalableimage}.
We observe that additional sampling steps have varying impacts on different metrics.
A single step maintains the highest similarity in low-level details between the encoded and reconstructed images as measured by PSNR.
Increasing the number of steps deteriorates PSNR but improves the rFID and LPIPS metrics.
Using the $t$-spacing scheduler from~\citet{flow_to_the_mode_sargent2025flowmodemodeseekingdiffusion} with $\rho=2$ or $\rho=4$, perceptual-level distortion measured by LPIPS reaches an optimal value around $6$ steps, while distribution-level metric rFID is at its lowest at $8$ steps.
We also see that this remains consistent across different model sizes.
As we aim to provide a generative-oriented decoder, this confirms our choice of 8 sampling steps for most of our models.
We additionally compare to a linear scheduler ($\rho=1$), using \Ssdd{}-M and a \Dito{}-XL model trained with LPIPS.
We observe a similar effect of increasing sampling steps happening at a slower pace.
This confirms that this effect doesn't come from our specific model or sampling choices but from the perceptual regularization shared by \Ssdd{}, \evae{} and LPIPS-regularized instances of \Dito{}.
As such, setting  the number of sampling steps selects the behavior of the decoder relative to the different distortion and distribution shift measures, and distillation  consolidates the selected behavior into a single-step model.

\mypar{Effect of sampling on fidelity and diversity}
To better understand the behavior behind the shift of model behavior displayed in \Cref{fig:sampling}, we measure the Density (fidelity metric) and Coverage (diversity metric)~\citep{density_coverage_pmlr-v119-naeem20a} between the ground truth and reconstructed set on ImageNet \res{256}, when varying the number of sampling steps.
We evaluate models with diverse sizes (S, M, L) and samplers ($\rho=0,2,4$), and display the results in~\Cref{fig:density_coverage}.
In each case, both fidelity metrics (FID and Density) drop around the same point.
The Coverage stays high even with a high number of sampling steps and does not suffer from meaningful drops.
We hypothesize that LPIPS-regularized diffusion decoders suffer from an \textit{overshooting} issue, where the diversity increases outside the bounds of the training distribution support after enough steps.

\begin{figure*}[bt]
\centering
    \begin{subfigure}{.5\textwidth}
      \centering
        \includegraphics[width=\textwidth]{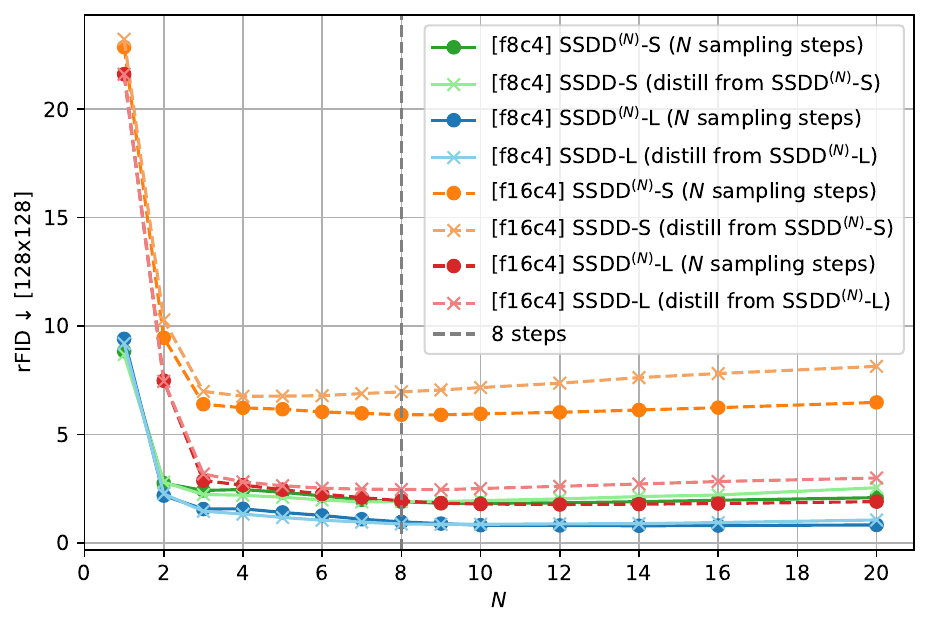}
        \caption{Reconstruction on ImageNet 256x256.}
        \vspace{1.5em}
        \label{fig:rgfid_steps_rfid}
        \tabfigendspace
    \end{subfigure}
    \begin{subfigure}{.5\textwidth}
      \centering
        \includegraphics[width=\textwidth]{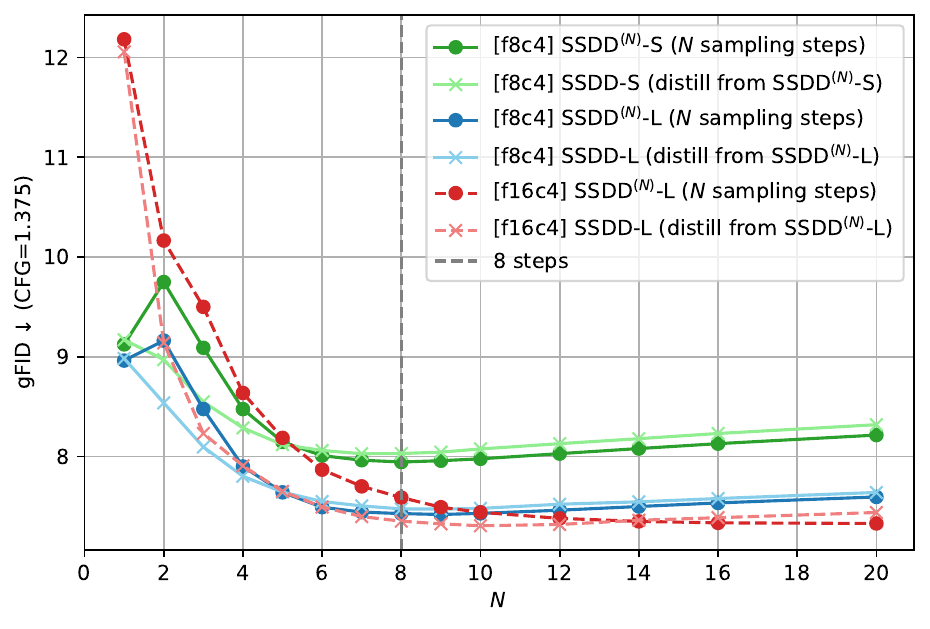}
        \caption{Generation using a DiT-XL/2 trained for 400k steps on ImageNet 256x256, using CFG.}
        \vspace{0.5em}
        \label{fig:rgfid_steps_gfid}
        \tabfigendspace
    \end{subfigure}
    \caption{
        \textbf{Effect of the number of sampling steps on reconstruction and generation metrics, evaluated on ImageNet \res{256}.}
        For distilled models, $N$ is the number of sampling steps of the teacher model.
        Students are distilled for only $5K$ steps. 
    }
    \label{fig:rgfid_steps}
\end{figure*}

\mypar{Effect of distillation teacher steps on reconstruction and generation}
We analyze in \Cref{fig:rgfid_steps} the behavior of both teacher and distilled student models on reconstruction and generation tasks, depending on the number $N$ of sampling steps.
We observe in \Cref{fig:rgfid_steps_rfid} that the behavior of the distilled models follows closely the teacher on reconstruction tasks when using a low compression rate of f8c4, both for a small (S) and large (L) model.
At a higher rate (f16c4), we observe a slight degradation of performances.
On the contrary, for generation in \Cref{fig:rgfid_steps_gfid} the student models display improvements over the teacher models for low number of reference sampling steps, with similar (at f8c4) or better results (at f16c4) with 8 teacher steps.
This shows that our distillation method preserves most model capabilities for both reconstruction and generative tasks.

\input{tables/distillation}

\mypar{Impact of distillation on final model quality}
We control for the impact of single-step distillation on final reconstruction and evaluation results at resolution \res{256}  in \Cref{tab:distillation}.
We follow the distillation procedure from \Cref{subsec:sampling_distill} (details in \Cref{app:implementation_details}).
We evaluate the reconstruction distribution shift (rFID) and perceptual distortion (DreamSim) and the impact on generative models by decoding from a DiT-XL/2.
While distillation caused a noticeable impact on the \res{128} \Ssdd{}-M model (see \Cref{tab:ablation_from_dito}),
we observe slight to no effect at \res{256} models, with single-step distilled \Ssdd{} using f16c4 encoder even outperforming teacher models on generative tasks.
Moving to the higher compression ratio of f16c4, we observe a higher impact on the reconstruction FID, but a reduced impact on generative FID and DreamSim metrics.
We conclude that, while distillation can hinder distribution-shift metrics on high compression ratios (f16c4, or f8c4 with smaller images), it has little effect on perceptual quality (DreamSim) and generative applications (gFID), while providing important speed-ups.

\section{Additional results}
\label{app:additional_results}

\input{tables/from_pretrained_encoders}

\mypar{Reconstruction from existing encoders}
To evaluate whether \Ssdd{} can be used as a replacement for existing decoders for existing encoders, or whether it needs specialized jointly trained encoders, we train our model on top of existing frozen encoders with varying compression ratios.
On the f8c4 compression ratio, we train decoders on top of SD-VAE~\citep{sd1_Rombach_2022_CVPR} and \Dito{}-XL-LPIPS~\citep{dito_chen2025diffusionautoencodersscalableimage} encoders.
To evaluate high spatial downsampling, we use DC-AE~\citep{dcae_chen2025deep} (f32c32 and f64c128).
We adapt \Ssdd{} conditioning to the DC-AE setting by adding a linear projection at the input, mapping each input token (with channel dimension $32$ or $128$) to a feature map of size $32\times32\times32$ or $64\times64\times128$.

We show in \Cref{tab:from_pretrained_encoders} that \Ssdd{} outperforms the original decoders on reconstruction performance across all encoders and compression ratios, despite being conditioned on features optimized for a different architecture.
This demonstrates that \Ssdd{} is a versatile decoder model that can be used to improve the reconstruction quality from existing auto-encoders.

\begin{figure*}[bt]
\centering
    \begin{subfigure}{.5\textwidth}
      \centering
        \includegraphics[width=\textwidth]{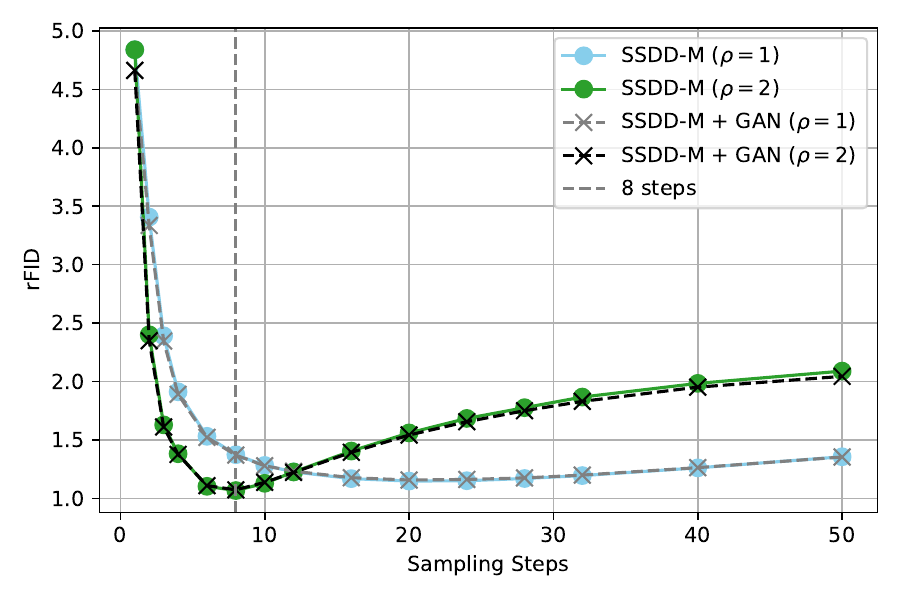}
        \label{fig:gan_sampling_1}
        \tabfigendspace
    \end{subfigure}
    \begin{subfigure}{.5\textwidth}
      \centering
        \includegraphics[width=\textwidth]{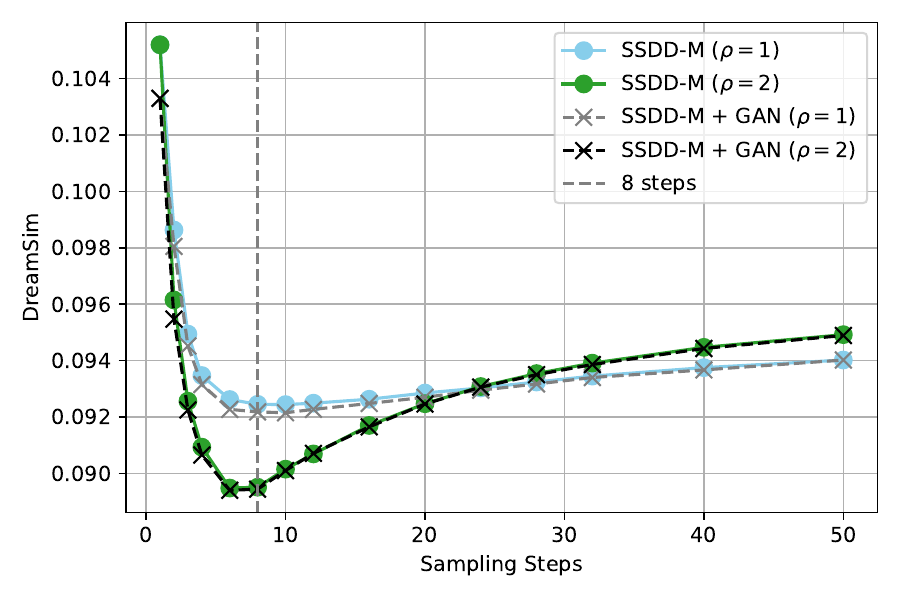}
        \label{fig:gan_sampling_2}
        \tabfigendspace
    \end{subfigure}
    \caption{
        \textbf{GAN sampling.}
        Evolution of the metrics depending on the number of steps $N$.
        Models directly trained at $128\!\times\!128$.
        Evaluated on ImageNet $128\!\times\!128$.
    }
    \label{fig:gan_sampling}
\end{figure*}

\mypar{Impact of GAN loss on sampling}
According to~\citet{evae_zhao2025epsilonvaedenoisingvisualdecoding}, the fact that a low number of sampling steps is optimal for \evae{} results from the \textit{denoising trajectory matching}, referring to their adapted GAN loss.
This is related to similar work that introduces an adversarial loss on latent diffusion models~\citep{xiao2022tackling}.
We  test this hypothesis by using the same adversarial loss as~\citet{evae_zhao2025epsilonvaedenoisingvisualdecoding} on top of our model, and sweeping over gan loss weights to select optimal one.
We train a single \Ssdd{} with an f8c4 encoder at resolution $128\!\times\!128$, and evaluate with a varying number of steps with linear ($\rho=0$) and shifted ($\rho=2$) schedulers.
Results are displayed in \Cref{fig:gan_sampling}.
The adversarial loss has negligible impact on both the rFID and DreamSim metrics, yielding very slight improvements at some specific points.
Additionally, the optimal number of sampling steps for both metrics is not shifted, as the curves keep the same shape.
We conclude that the LPIPS loss is the main driver of this behavior, which we also observed in other GAN-free models in \Cref{fig:sampling}.

\input{tables/app_fine_tuning}

\mypar{Fine-tuning and generalization at higher resolutions}
To evaluate the impact of target-resolution fine-tuning of our base model trained with multi-resolution data augmentation, we evaluate base and fine-tuned \Ssdd{} models at varying resolutions and model sizes, and display the results in \Cref{tab:fine_tuning}.
We observe that for almost all settings and metrics, the base model ranks second and already achieves high-quality results.
This ensures that it provides a high-quality shared pre-training for all resolutions with a low training cost.

\input{tables/app_extended_ablations}

\mypar{Extended distillation ablations: GAN loss and L2-only distillation}
\label{app:extended_ablations}
We extend the ablation of \Cref{tab:ablation_from_dito} with two further controls on the single-step \Ssdd-M distilled decoder (f8c4, \res{128}), reported in \Cref{tab:app_extended_ablations}.
First, adding the adversarial denoising trajectory matching loss from~\citet{evae_zhao2025epsilonvaedenoisingvisualdecoding} on top of distillation, with the GAN weight tuned for best rFID, only improves rFID by $0.01$. This very small gain does not warrant the added training complexity and instability of adversarial training, and is consistent with our earlier observation on the multi-step teacher (\Cref{tab:ablation_from_dito}, last row).
Second, replacing our loss-preserving distillation (flow-matching and LPIPS computed against teacher outputs) by an L2-regression distillation as in~\citet{luhman2021knowledgedistillationiterativegenerative} degrades the distilled rFID by $0.16$, confirming that in our setting preserving the perceptual term during distillation is necessary to retain the perceptual quality of the teacher in a single step.

\mypar{Comparison with single-stage MeanFlow training}
\label{app:meanflow}
A natural alternative to our two-stage \emph{train-then-distill} pipeline is to train a single-step decoder directly with MeanFlow~\citep{geng2025mean}.
We compare both approaches in the setting of \Cref{tab:ablation_from_dito} (f8c4, \res{128}).
In terms of training cost, MeanFlow is approximately $2\times$ slower per step than our flow-matching pre-training, because each gradient evaluation requires a Jacobian-vector product through the network.
In terms of quality, multi-step MeanFlow with its optimal sampling schedule already degrades by $+0.62$ rFID (and improves PSNR by $+0.6$) compared to our distilled \Ssdd-M, and single-step MeanFlow further trades $+6.7$ rFID for $+1.6$ PSNR, recovering the same perception--distortion movement we control through the choice of teacher steps in \Cref{subsec:sampling_distill}.
Beyond the cost difference, the two-stage pipeline also \emph{decouples} the selection of a perception--distortion operating point (set by the multi-step teacher schedule) from the single-step inference efficiency (set by distillation), which is not straightforward to do with a one-stage training objective.

\mypar{Component contribution isolation}
\label{app:component_isolation}
We complement the sequential ablation of \Cref{tab:ablation_from_dito} with a grouping of components by their role in the final method.
Two components are \emph{essential} to the perceptual quality of the decoder: replacing the \Dito{}-S backbone by the \Ssdd-M decoder ($-1.24$ rFID over the +LPIPS reference, $-1.16$ over \Dito{}-S-LPIPS in the table) and adding the REPA loss ($-0.68$ rFID over the +LPIPS reference, $-0.44$ in the table).
The next four components --- KL-regularization on the encoder, logit-normal noise sampling, $t$-spacing scheduler, and EMA --- are \emph{standard recipe improvements} ($-0.06$ to $-0.18$ rFID each, cumulatively about $-0.5$ rFID) that could be swapped with similar techniques without changing the conclusions.
Finally, the shared encoder, shared pre-training, and single-step distillation are \emph{not necessary} to obtain a strong decoder: their role is to enable a single latent space shared by all decoder sizes, to reduce per-decoder training cost, and to reach single-step inference. They are pipeline choices, not quality enablers.

\section{Qualitative results}
\label{app:qualitative_results}

\begin{figure*}[hbtp]
    \centering
    \includegraphics[width=\textwidth]{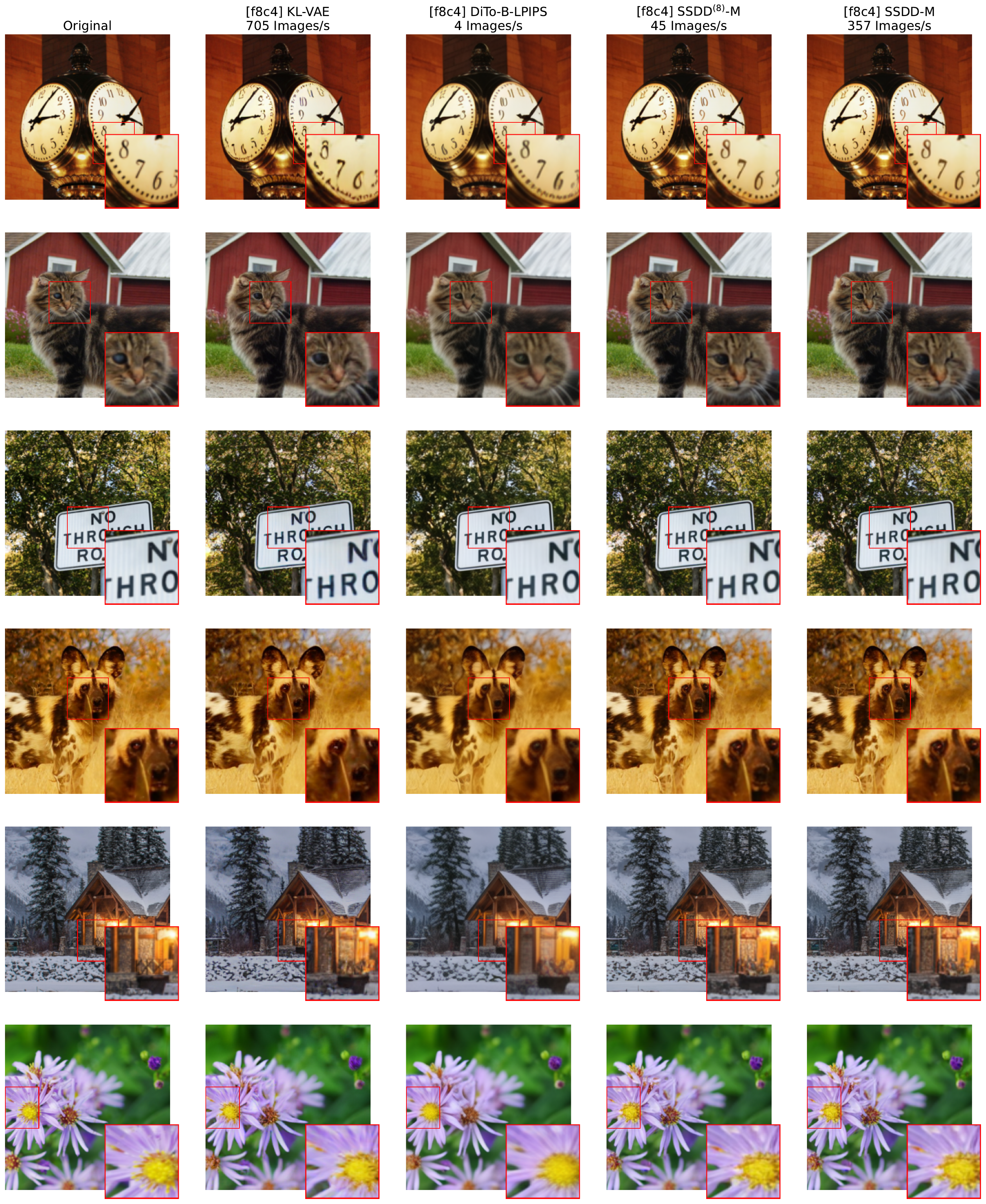}
    \figmidspace
    \caption{
        \textbf{Qualitative comparison between decoders.}
        Input images are of resolution $256\times 256$ and compressed into a $16\times 16\times 4$ latent representation by an f8c4 encoder.
        Left to right: original image, KL-VAE (47.2M parameters), \Dito{}-B (155.2M parameters, 20 sampling steps), \ssddmulti{}$^{(8)}$-M (48.0M, 8 sampling steps), and \Ssdd{}-M (48.0M, distilled into a single sampling step).
    }
    \label{fig:qual_f8c4}
    \tabfigendspace
\end{figure*}

\begin{figure*}[hbtp]
    \centering
    \includegraphics[width=\textwidth]{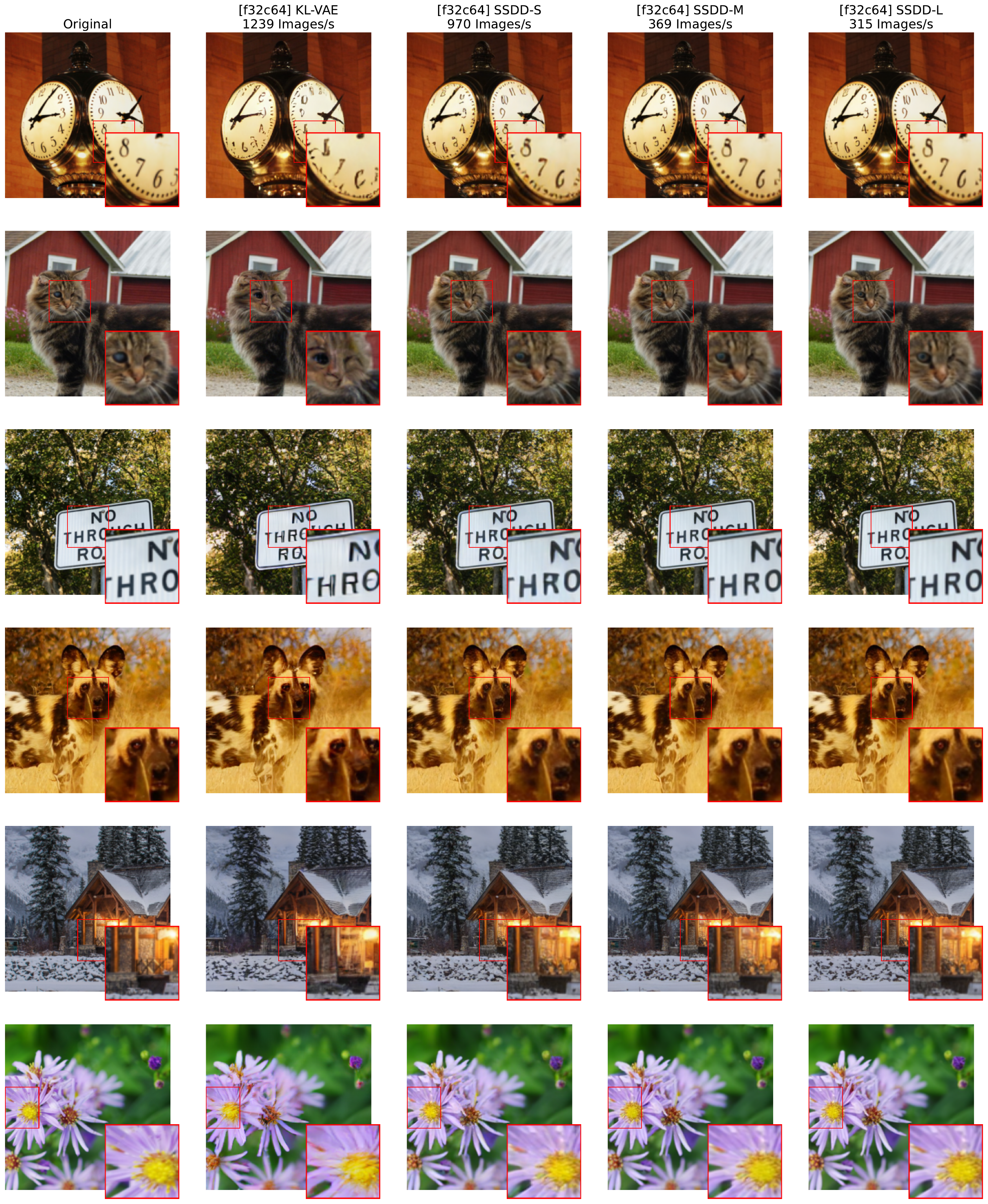}
    \caption{
        \textbf{Qualitative comparison using a deep encoder.}
        Input images are of resolution $256\times 256$, and compressed into a $4\times 4\times 64$ latent representation by an f32c64 encoder.
        Left to right: original image, KL-VAE (47.2M parameters), and single-step distilled models \Ssdd-S (13.4M),  \Ssdd-M (48.0M) and \Ssdd-L (85.2M).
    }
    \label{fig:qual_f32c64}
    \tabfigendspace
\end{figure*}

We provide a visual comparison of both distilled and non-distilled \Ssdd{} with deterministic (KL-VAE) and generative (\Dito{}) decoders in \Cref{fig:qual_f8c4,fig:qual_f32c64}.
The same random noise is used to generate each reconstruction for each image, ensuring alignment between distilled and non-distilled model behavior.

In \Cref{fig:qual_f8c4}, using a low spatial downsampling (f8c4), KL-VAE generates distorted details, and \Dito{} outputs blurry low-level features (see for example the clock image).
On the same images, \Ssdd{}$^{(8)}$-M generates sharp, low-distortion, and realistic details.
Comparing the multi-step and distilled \Ssdd{} models, conditioned on the same latent variable $z$ and the same noise $\varepsilon$, their outputs are almost indistinguishable, with the same distortion of low-level features.

In \Cref{fig:qual_f32c64}, using a deeper encoder, the reconstruction quality of KL-VAE drops severely, generating non-realistic images (see for example the cat image, and the text on the sign).
Our fastest decoder, \Ssdd{}-S, still generates similarly sharp details, while remaining in the range of realistic images.
Additionally, larger models (\Ssdd{}-M and \Ssdd{}-L) output visibly sharper and more accurate reconstructions.

\begin{figure*}[hbtp]
    \centering
    \includegraphics[width=.97\textwidth]{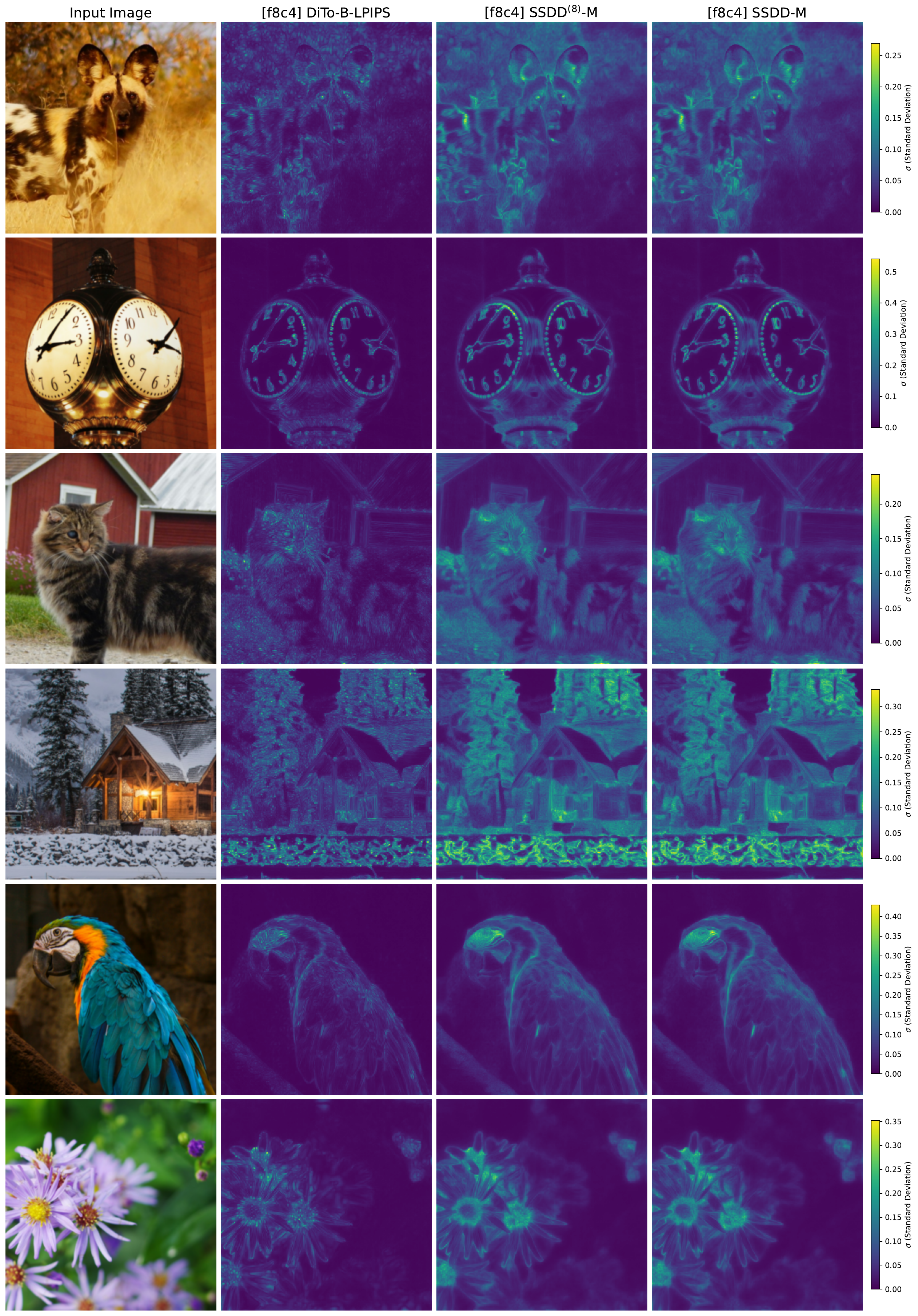}
    \figmidspace
    \caption{
        \textbf{Diversity maps of reconstructed images.}
        Left to right: original image, DiTo-B-LPIPS (155.2M parameters, 50 sampling steps), \ssddmulti{}$^{(8)}$-M (48.0M, 8 sampling steps) and \Ssdd{}-M (48.0M, distilled into a single sampling step).
        We compute the standard deviation for each pixel over $64$ decoding processes for each image.
        Each row uses a single shared scale to display diversity maps.
    }
    \label{fig:map_diversity}
    \tabfigendspace
\end{figure*}

We provide a visualization in \Cref{fig:map_diversity} of the diversity of images reconstructions by \Dito{} and both multi-step and distilled \Ssdd{}.
Diversity is not uniformly distributed and is more prominent around details (eyes, numbers, dots) and edges.
Both \Dito{} and \Ssdd{} diversity maps are focused around the same regions, with smoother and slightly larger areas for the latter.
\Ssdd{} also benefits from higher diversity levels.
The distilled \Ssdd{} decoder has similarly diverse generations as the teacher, illustrating how distillation preserves both quality (\Cref{fig:qual_f8c4}) and diversity (\Cref{fig:map_diversity}).

\begin{figure*}[hbtp]
    \centering
    \includegraphics[width=\textwidth]{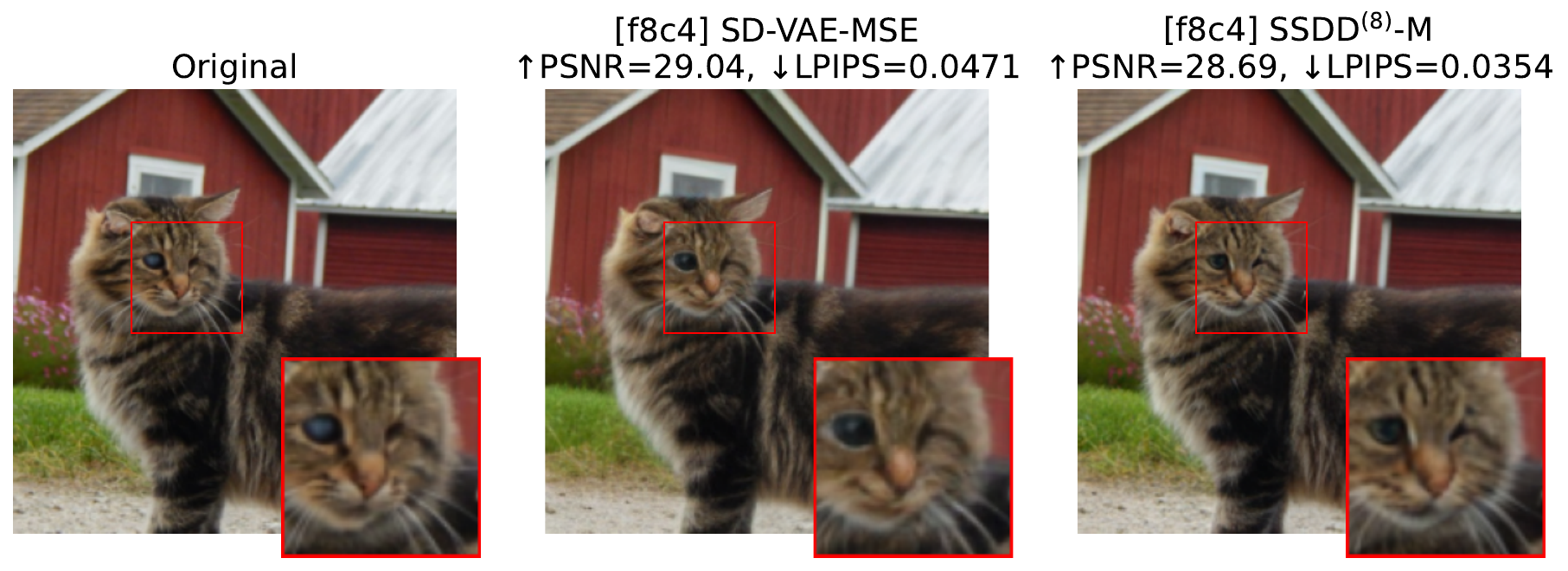}
    \figmidspace
    \caption{
        \textbf{Perception--distortion trade-off in reconstructions.}
        Comparison of \sdvae{} and \Ssdd{}-M (f8c4, \res{256}) on representative ImageNet samples.
        \Ssdd{} obtains better perceptual results (lower LPIPS) at the cost of a lower PSNR, with sharper and more realistic details that may differ from the original; \sdvae{} produces more blurry results favored by PSNR.
    }
    \label{fig:qual_tradeoff}
    \tabfigendspace
\end{figure*}

\Cref{fig:qual_tradeoff} provides an additional side-by-side comparison between \sdvae{} and \Ssdd{} on the f8c4 setting, illustrating the perception-distortion trade-off discussed in \Cref{sec:intro,app:theory}: \Ssdd{} produces sharper and more realistic high-frequency details, favored by perceptual metrics (LPIPS, rFID), while \sdvae{} yields blurrier reconstructions that score higher on pixel-level distortion metrics (PSNR, SSIM). The two crops at the bottom show that \Ssdd{} details may differ from the original on highly ambiguous regions, which is expected for a generative decoder.

%% file: tables/model_configs.tex
\begin{table}[hbtp]
\caption{
    Architecture configurations of \Ssdd{} decoders.
}
\label{tab:model_configs}
\begin{center}
\scriptsize

\begin{tabular}{lrrrr}
\toprule

\textbf{Model} & \textbf{\#D} & \textbf{Channels} & \textbf{Depth multipliers} & \textbf{\# Blocks} \\
\midrule
\Ssdd{}-S & 13.4M & $48$ & $\{1, 2, 3, 3\}$ & $8$ \\
\Ssdd{}-B & 20.2M & $64$ & $\{1, 2, 3, 3\}$ & $10$ \\
\Ssdd{}-M & 48.0M & $96$ & $\{1, 2, 3, 3\}$ & $12$ \\
\Ssdd{}-L & 85.2M & $96$ & $\{1, 2, 4, 4\}$ & $16$ \\
\Ssdd{}-XL & 153.8M & $128$ & $\{1, 2, 4, 4\}$ & $16$

 \\ \bottomrule
\end{tabular}

\end{center}
\end{table}

%% file: tables/distillation.tex
\newcommand{\disgood}[1]{\textit{\color{OliveGreen}#1}}
\newcommand{\disbaad}[1]{\textit{\color{Bittersweet}+#1}}
\newcommand{\disworse}[1]{\textit{\color{Red}+#1}}
\newcommand{\disneut}[1]{\textit{\color{Gray}±#1}}

\begin{table}[bt]
\caption{
    {\bf Impact of distillation on  reconstruction and generation.} 
    The distilled model has a $8\times$ higher throughput for decoding.
    Results on ImageNet $256\!\times\!256$, using and 50k distillation steps.
    Generation results use a DiT-XL/2 trained for 400k steps and a CFG of 1.375.
}
\label{tab:distillation}
\begin{center}
\scriptsize

\begin{tabular}{clcccrlrlrl}
\toprule

\multicolumn{2}{c}{} & \multicolumn{3}{c}{\textbf{8-steps sampling}} & \multicolumn{6}{c}{\textbf{Single-step distilled model}}                          \\
\cmidrule(lr){3-5} \cmidrule(lr){6-11}

&   &
  \textbf{rFID\odec} &
  \textbf{DreamSim\odec} &
  \multicolumn{1}{c}{\textbf{gFID\odec}} &
  \multicolumn{2}{l}{\textbf{rFID\odec}} &
  \multicolumn{2}{l}{\textbf{DreamSim\odec}} &
  \multicolumn{2}{l}{\textbf{gFID\odec}} \\

\midrule

\multirow{5}{*}{\rotatebox{90}{\textbf{f8c4}}}

        & \ssddone{-S} & 0.46 & 0.039 & 7.95 & 0.46 & \disneut{0.00} & 0.039 & \disneut{0.000} & 8.02 & \disbaad{0.07} \\
        & \ssddone{-B} & 0.42 & 0.037 & 7.80 & 0.42 & \disneut{0.00} & 0.036 & \disgood{-0.001} & 7.85 & \disbaad{0.05} \\
        & \ssddone{-M} & 0.36 & 0.034 & 7.58 & 0.39 & \disbaad{0.03} & 0.034 & \disneut{0.000} & 7.63 & \disbaad{0.05} \\
        & \ssddone{-L} & 0.34 & 0.032 & 7.43 & 0.36 & \disbaad{0.02} & 0.032 & \disneut{0.000} & 7.48 & \disbaad{0.05} \\
        & \ssddone{-XL} & 0.33 & 0.031 & 7.43 & 0.35 & \disbaad{0.02} & 0.031 & \disneut{0.000} & 7.46 & \disbaad{0.03} \\

\midrule

\multirow{5}{*}{\rotatebox{90}{\textbf{f16c4}}}

        & \ssddone{-S} & 2.14 & 0.122 & 9.88 & 2.29 & \disworse{0.15} & 0.121 & \disgood{-0.001} & 9.83 & \disgood{-0.05} \\
        & \ssddone{-B} & 1.65 & 0.112 & 9.13 & 1.77 & \disworse{0.12} & 0.111 & \disgood{-0.001} & 8.98 & \disgood{-0.15} \\
        & \ssddone{-M} & 1.05 & 0.096 & 8.15 & 1.23 & \disworse{0.18} & 0.096 & \disneut{0.000} & 7.99 & \disgood{-0.16} \\
        & \ssddone{-L} & 0.84 & 0.089 & 7.59 & 0.96 & \disworse{0.12} & 0.089 & \disneut{0.000} & 7.34 & \disgood{-0.25} \\
        & \ssddone{-XL} & 0.78 & 0.084 & 7.40 & 0.89 & \disworse{0.11} & 0.083 & \disgood{-0.001} & 7.12 & \disgood{-0.28}

 \\ \bottomrule
\end{tabular}

\end{center}
\end{table}

%% file: tables/from_pretrained_encoders.tex
\begin{table}[hbtp]
\caption{
    \textbf{Reconstruction from frozen pre-trained encoders.}
    DITO-XL-LPIPS is trained without noise synchronization. 
}
\label{tab:from_pretrained_encoders}
\begin{center}
\scriptsize

\begin{tabular}{lcc|ccc}
\toprule

    & \textbf{Encoder} & \textbf{Decoder} & \textbf{rFID\odec} & \textbf{LPIPS\odec} & \textbf{DreamSim\odec} \\

\midrule

\multirow{10}{*}{\rotatebox{90}{\textbf{f8c4}}}
    & \multirow{5}{*}{\makecell{\sdvae{} \\ \citep{sd1_Rombach_2022_CVPR} }}         & \sdvae{}  & 0.69 & 0.061 & 0.040 \\
    &                                   & \ssddone{-S}         & 0.55 & 0.063 & 0.041 \\
    &                                   & \ssddone{-B}         & 0.49 & 0.061 & 0.039 \\
    &                                   & \ssddone{-M}         & 0.44 & 0.058 & 0.036 \\
    &                                   & \ssddone{-L}         & \textbf{0.39} & \textbf{0.056} & \textbf{0.034} \\

\cmidrule(l){2-6}

    & \multirow{5}{*}{\makecell{\Dito{}-XL-LPIPS (noise\_sync) \\ \citep{dito_chen2025diffusionautoencodersscalableimage}}} & \Dito{}-XL-LPIPS & 1.37 & 0.083 & 0.058 \\
    &                                   & \ssddone{-S}         & 0.91 & 0.074 & 0.052 \\
    &                                   & \ssddone{-B}         & 0.78 & 0.070 & 0.049 \\
    &                                   & \ssddone{-M}         & 0.64 & 0.068 & 0.045 \\
    &                                   & \ssddone{-L}         & \textbf{0.58} & \textbf{0.065} & \textbf{0.042} \\

\midrule

\multirow{5}{*}{\rotatebox{90}{\textbf{f32c32}}}
    & \multirow{5}{*}{\makecell{DC-AE \\ \citep{dcae_chen2025deep} }}         & DC-AE  & 0.69 & 0.081 & \textbf{0.045} \\
    &                                   & \ssddone{-S}         & 0.92 & 0.091 & 0.056 \\
    &                                   & \ssddone{-B}         & 0.82 & 0.088 & 0.053 \\
    &                                   & \ssddone{-M}         & 0.66 & 0.083 & 0.047 \\
    &                                   & \ssddone{-L}         & \textbf{0.60} & \textbf{0.080} & \textbf{0.045} \\

\midrule

\multirow{5}{*}{\rotatebox{90}{\textbf{f64c128}}}
    & \multirow{5}{*}{\makecell{DC-AE \\ \citep{dcae_chen2025deep} }}         & DC-AE  & 0.81 & 0.087 & 0.051 \\
    &                                   & \ssddone{-S}         & 0.97 & 0.094 & 0.058 \\
    &                                   & \ssddone{-B}         & 0.85 & 0.091 & 0.055 \\
    &                                   & \ssddone{-M}         & 0.70 & 0.085 & 0.049 \\
    &                                   & \ssddone{-L}         & \textbf{0.63} & \textbf{0.082} & \textbf{0.046} \\

\bottomrule
\end{tabular}

\end{center}
\end{table}

%% file: tables/app_fine_tuning.tex
\begin{table}[hbtp]
\caption{
    \textbf{Effect of fine-tuning at target resolution.}
    We highlight the \textbf{first} and \underline{second} best values of each metric for every evaluation setting.
}
\label{tab:fine_tuning}
\begin{center}
\scriptsize

\resizebox{\textwidth}{!}{
\begin{tabular}{lc|ccc|ccc|ccc}
\toprule

\multirow{2}{*}{\textbf{Model}} &
  \multirow{2}{*}{\textbf{Fine-tuned at}} &
  \multicolumn{3}{c}{\textbf{Evaluated at \res{128}}} &
  \multicolumn{3}{c}{\textbf{Evaluated at \res{256}}} &
  \multicolumn{3}{c}{\textbf{Evaluated at \res{512}}} \\
 &
   &
  \textbf{rFID} &
  \textbf{PSNR} &
  \textbf{DreamSim} &
  \textbf{rFID} &
  \textbf{PSNR} &
  \textbf{DreamSim} &
  \textbf{rFID} &
  \textbf{PSNR} &
  \textbf{DreamSim} \\

\midrule

\ssddmulti$^{(8)}${-S} & N/A (base model) & \underline{2.13} & \underline{22.92} & \underline{0.098} & \underline{0.57} & \underline{24.00} & \underline{0.041} & \underline{0.37} & \underline{25.67} & 0.036 \\
\ssddmulti$^{(8)}${-S} & 128x128 & \textbf{1.88} & \textbf{22.99} & \textbf{0.095} & \underline{0.57} & 23.93 & \underline{0.041} & \underline{0.37} & 25.58 & \underline{0.035} \\
\ssddmulti$^{(8)}${-S} & 256x256 & 2.36 & 22.84 & 0.102 & \textbf{0.47} & \textbf{24.08} & \textbf{0.040} & \textbf{0.25} & \textbf{25.91} & \textbf{0.031} \\

\midrule

\ssddmulti$^{(8)}${-B} & N/A (base model) & \underline{1.69} & \underline{23.01} & \underline{0.093} & \underline{0.48} & \underline{24.09} & \underline{0.039} & \underline{0.29} & \underline{25.81} & \underline{0.032} \\
\ssddmulti$^{(8)}${-B} & 128x128 & \textbf{1.50} & \textbf{23.11} & \textbf{0.091} & 0.51 & \underline{24.09} & \underline{0.039} & 0.31 & 25.57 & 0.035 \\
\ssddmulti$^{(8)}${-B} & 256x256 & 2.03 & 22.97 & 0.097 & \textbf{0.42} & \textbf{24.18} & \textbf{0.037} & \textbf{0.21} & \textbf{25.99} & \textbf{0.029} \\

\midrule

\ssddmulti$^{(8)}${-M} & N/A (base model) & \underline{1.28} & \underline{23.19} & \underline{0.088} & \underline{0.44} & 24.26 & \underline{0.037} & \underline{0.27} & \underline{25.91} & \underline{0.032} \\
\ssddmulti$^{(8)}${-M} & 128x128 & \textbf{1.17} & \textbf{23.29} & \textbf{0.086} & 0.47 & \underline{24.27} & \underline{0.037} & 0.30 & 25.80 & 0.034 \\
\ssddmulti$^{(8)}${-M} & 256x256 & 1.55 & 23.18 & 0.091 & \textbf{0.40} & \textbf{24.38} & \textbf{0.035} & \textbf{0.21} & \textbf{26.20} & \textbf{0.029} \\

\midrule

\ssddmulti$^{(8)}${-L} & N/A (base model) & \underline{1.10} & \underline{23.26} & \underline{0.085} & \underline{0.43} & 24.37 & \underline{0.035} & 0.26 & \underline{26.14} & \underline{0.031} \\
\ssddmulti$^{(8)}${-L} & 128x128 & \textbf{0.97} & \textbf{23.40} & \textbf{0.083} & \underline{0.43} & \underline{24.44} & \underline{0.035} & \underline{0.24} & 26.13 & \underline{0.031} \\
\ssddmulti$^{(8)}${-L} & 256x256 & 1.23 & 23.28 & 0.088 & \textbf{0.37} & \textbf{24.49} & \textbf{0.033} & \textbf{0.19} & \textbf{26.35} & \textbf{0.028} \\

\midrule

\ssddmulti$^{(8)}${-XL} & N/A (base model) & \underline{1.02} & 23.41 & \underline{0.083} & \underline{0.42} & \underline{24.43} & \underline{0.034} & \underline{0.24} & \underline{26.10} & \underline{0.031} \\
\ssddmulti$^{(8)}${-XL} & 128x128 & \textbf{0.94} & \textbf{23.50} & \textbf{0.081} & 0.46 & 24.42 & 0.035 & 0.25 & 25.95 & 0.032 \\
\ssddmulti$^{(8)}${-XL} & 256x256 & 1.07 & \underline{23.43} & 0.086 & \textbf{0.38} & \textbf{24.60} & \textbf{0.033} & \textbf{0.20} & \textbf{26.40} & \textbf{0.029}

 \\ \bottomrule
\end{tabular}
}

\end{center}
\end{table}

%% file: tables/app_extended_ablations.tex
\begin{table}[t]
\caption{
    \textbf{Extended ablations on f8c4 \res{128}, complementing \Cref{tab:ablation_from_dito}.}
    \emph{+ GAN loss (single-step)}: adding the \evae{} adversarial denoising trajectory matching loss on top of the distilled single-step \Ssdd{}-M, with GAN weighting tuned for best rFID.
    \emph{L2-only distillation}: replacing our loss-preserving distillation (flow-matching + LPIPS computed against teacher outputs) by an L2 regression on teacher outputs as in~\citet{luhman2021knowledgedistillationiterativegenerative}.
}
\label{tab:app_extended_ablations}
\begin{center}
\scriptsize

\begin{tabular}{l|rlrlrl|c}
\toprule
\textbf{Ablation} & 
\multicolumn{2}{l}{\textbf{rFID\odec} } & 
\multicolumn{2}{l}{\textbf{PSNR\oinc}}& 
\multicolumn{2}{l|}{\textbf{DreamSim\odec}}&
$N$\\
\midrule
\Ssdd-M ($\S$, single-step distilled)   & 1.04 &                & 23.28 &                & 0.084 &                & 1 \\
($\S$) + GAN loss (tuned weighting)     & 1.03 & \abgood{-0.01} & 23.32 & \abgood{+0.04} & 0.084 & \abneut{0.000} & 1 \\
($\S$) + L2-only distillation             & 1.20 & \abbaad{+0.16} & 23.45 & \abgood{+0.17} & 0.092 & \abbaad{+0.008} & 1 \\
\bottomrule
\end{tabular}
\end{center}
\end{table}